\pdfoutput=1

\documentclass[11pt]{article}

\usepackage[]{ACL2023}

\usepackage{times}
\usepackage{latexsym}

\usepackage[T1]{fontenc}

\usepackage[utf8]{inputenc}

\usepackage{microtype}

\usepackage{multirow}
\usepackage{booktabs}
\usepackage{xcolor}
\usepackage{adjustbox}						
\usepackage{fancybox}
\usepackage{ulem}
\usepackage{wrapfig,lipsum,booktabs}
\usepackage{amsmath}
\usepackage{array}
\usepackage{makecell}
\usepackage{amsfonts}
\usepackage{footnote}
\usepackage{boxedminipage}

\usepackage[colorinlistoftodos]{todonotes}
\usepackage{xurl}

\newcommand{\ourmodel}{\texttt{ALERT}}

\newcommand{\cb}[1]{\mbox{\CheckBox[bordercolor=,name=#1,height=0.4cm,width=0.4cm]{}{\ }}}
\definecolor{tblue}{HTML}{174992}
\definecolor{tyellow}{HTML}{e7cd79}
\definecolor{trose}{HTML}{bc6880}
\definecolor{torange}{HTML}{e39e36}
\definecolor{lightyellow}{HTML}{fff2cc}
\definecolor{lightgreen}{HTML}{e1efdd}

\newcommand{\RN}[1]{%
	\textup{\lowercase\expandafter{\it \romannumeral#1}}%
}

\usepackage{inconsolata}

\title{\ourmodel: Adapting Language Models to Reasoning Tasks}

\author{Ping Yu${ }^{\spadesuit}$ \quad
Tianlu Wang${ }^{\spadesuit}$ \quad
Olga Golovneva${ }^{\spadesuit}$ \quad
Badr AlKhamissi${ }^{\triangle}$  \\
\textbf{Siddharth Verma${ }^{\triangle}$ \enskip
Zhijing Jin${ }^{\ddagger\triangle}$ \enskip
Gargi Ghosh${ }^{\spadesuit}$ \enskip
Mona Diab${ }^{\spadesuit}$ \enskip
Asli Celikyilmaz${ }^{\spadesuit}$ } \\
${ }^{\spadesuit}$Meta AI \quad
${ }^{\triangle}$Work done at Meta AI\\
${ }^{\ddagger}$Max Planck Institute \& ETH\\
\{pingyu,aslic\}@meta.com}

\begin{document}
\maketitle
\begin{abstract}
Recent advancements in large language models have enabled them to perform well on complex tasks that require step-by-step reasoning with few-shot learning. 
However, it is unclear whether these models are applying reasoning skills they have learned during pre-training, or if they are simply memorizing their training corpus at finer granularity and have learned to better understand their context.
To address this question, we introduce~\ourmodel, a benchmark and suite of analyses for evaluating reasoning skills of language models. \ourmodel~enables comparing pre-trained and finetuned models on complex tasks that require reasoning skills to solve them. Our benchmark provides a test bed to assess any language model on fine-grained reasoning skills, which spans over 20 datasets and covers 10 different reasoning skills. To prove the efficacy of \ourmodel~we  investigate \textit{the role of finetuning}. Our extensive empirical analysis shows that language models acquire reasoning skills such as textual entailment, abductive reasoning, and analogical reasoning during the finetuning stage compared to pretraining stage. Another finding is when language models are finetuned they tend to overfit to the prompt template, which hurts the robustness of models resulting in generalization problems. 
\end{abstract}

\section{Introduction}
\label{sec:intro}

Large language models (LLMs) (e.g., GPT-3~\cite{gpt3}, PALM~\cite{palm2022}, OPT~\cite{opt}) have shown increasing in-context learning capabilities with scaling up the model and data sizes.
Despite this progress, even the largest of these models still struggle with tasks such as commonsense reasoning~\citep{west-etal-2022-symbolic}, and math word problems~\citep{hendrycksmath2021} which require arithmetic reasoning or symbolic manipulation~\citep{rytting2021leveraging}. 
Table~\ref{tab:reasoning_examples} presents some examples that require certain reasoning skills. Even the powerful LLMs (such as \textit{text-davinci-003}\footnote{\scriptsize{\url{https://beta.openai.com/docs/models/gpt-3}.}} and ChatGPT\footnote{\scriptsize{\url{https://chat.openai.com/chat}.}}) fail to make correct predictions.

\begin{table}[t]
    \centering
     \scalebox{0.72}{
    \begin{tabular}{@{}p{290pt}}
     \toprule
    The cafeteria had 23 apples. If they used 20 to make lunch and bought 6 more, how many apples do they have? \par
    \textbf{The answer is }\colorbox{red!30}{29 apples}.\\  
    \midrule
    Select the best translation into predicate logic. David teaches Chris. (c: Chris; d: David; Txy: x teaches y)
    (A) Tdc; (B) Tcd; (C) Tcc; (D) dTc. \textbf{The answer is} \colorbox{red!30}{(B) Tcd}. \\
    \midrule
    Isabella entered the hall. Olivia entered the hall. The apple is in the  blue$\_$treasure$\_$chest. Olivia exited the hall. Isabella moved the apple to the green$\_$basket.
    Question: Where does Isabella think that Olivia searches for the apple?  \textbf{The answer is} \colorbox{red!30}{Isabella thinks that Olivia searches for the apple in the} \colorbox{red!30}{green$\_$basket}.\\
     \bottomrule
    \end{tabular}
    }
    \caption{\footnotesize Examples from tasks that require reasoning skills and generated outputs from GPT-3 series \textit{text-davinci-003} engine. The failed outputs are highlighted in \colorbox{red!30}{red}. Predictions by ChatGPT are shown in Table~\ref{tab:app_reasoning_examples} in Appendix.}
    \label{tab:reasoning_examples}
\end{table}

To improve large LLMs' performance on tasks that require multiple steps of reasoning, recent work used different prompting methods which included a rationale with the final answer in the form of: scratchpad for arithmetic and logical reasoning~\citep{nye2021show}, chain-of-thought (CoT)~\citep{cot} for practically any tasks, or adding \textit{let's think step-by-step} ~\citep{kojima2022large} to prompt models to generate explanations.
Other works such as \citet{chung2022scaling} integrated step-by-step explanations into the finetuning stage (CoT-finetuning).
While these techniques may improve the accuracy and interpretability, it is not well understood which reasoning skills they rely on or to what degree they require higher-order reasoning. It is also uncertain how frequently the stated reasoning steps actually contribute to the final task predictions.
For instance, to correctly answer the questions in Table~\ref{tab:reasoning_examples} a combination of logical, commonsense, math and spatial reasoning skills are required. 

In this work, to gain a deeper understanding of LLMs reasoning abilities in in-context learning settings, 
we introduce \ourmodel, a new pipeline to benchmark different LLMs on various reasoning skills and provide analysis to assess reasoning abilities. 
Unlike existing commonly used benchmarks (e.g.,~\citet{niv1,niv2,srivastava2022beyond}), \ourmodel~can evaluate LLMs' fine-grained reasoning skills. It spans over 20 datasets and covers 10 different reasoning skills including logical, causal, commonsense, abductive, spatial, analogical, argument and deductive reasoning as well as textual entailment, and mathematics (see Figure~\ref{fig: teaser}). \ourmodel~enables easy benchmarking of any LM (e.g., pre-trained, finetuned, CoT-finetuned) on a rich set of new inference methods including zero-shot, few-shot and CoT.

Using \ourmodel,~we further investigate whether finetuning
can improve LMs' performance on downstream reasoning tasks. Specifically, we are interested in diagnosing what actually improved when we observe a performance increase on reasoning tasks. Is it because models have seen similar data in the finetuning stage? Or is it because models have seen prompts in a specific template and memorize the template during finetuning such as definitions provided in the NIV2 benchmark \cite{niv2}?
Or does the LLM actually acquired the required reasoning skill? We investigate these three possibilities.


To study the above questions, we compare three different model types (as shown in Figure~\ref{fig:baselines}): a pre-trained model and two types of finetuned models. Specifically:
\begin{itemize}
    \item \textbf{OPT} \cite{opt}: A baseline LLM a pre-trained model with no finetuning (figure (A) in Figure~\ref{fig:baselines});
    \item \textbf{OPT-FT}: Meta-finetuned OPT on reference answers \textit{without} explanations, illustrated in  
    (figure (B) in Figure~\ref{fig:baselines});
    \item \textbf{OPT-CoT}: Meta-finetuned OPT on data with rationales (explanations) \cite{chung2022scaling,alkhamissi2023opt} (figure (C) in Figure~\ref{fig:baselines}). 
\end{itemize}

Using these three types of models, we investigate \textit{the role of finetuning} on three dimensions:


\noindent\textbf{(1) Data memorization}: We investigate whether the performance improvements obtained after finetuning can be attributed to using similar or sometimes the exact same data as in the evaluation datasets.
To this end, we use vocabulary overlap to measure the extent to which the evaluation data is different from the finetuning data, i.e. We investigate whether the improvement is more significant when evaluation data and finetuning data are more similar.


\noindent\textbf{(2) Reasoning skills transfer:}
We investigate if certain reasoning skills can be more successfully permeated in LLMs than other reasoning skills. 
To verify this, we carefully divide the evaluation datasets into groups which require different reasoning skills. 
We compile held-out datasets as shown in Figure~\ref{fig: teaser} which require skills held-out from any of the 
 training datasets. 
This way, we expect to see larger improvements on in-domain skills compared to held-out skills if reasoning skills can be transferred during finetuning stages.


\noindent\textbf{(3) Prompt template memorization:} Our third hypothesis is that LLMs can overfit to data format used in the finetuning datasets such as training data format used in Figure~\ref{fig:baselines}. In other words, the consistency in data format helps LLMs better understand the instruction which then yields better performance after finetuning. To test this, we evaluate finetuned LLMs on datasets with 5 different prompt templates.

\begin{table}[t]
    \centering
    \scalebox{0.75}
	    {
    \begin{tabular}{lp{180pt}}
        \toprule
         \textbf{Reasoning Skills} & \textbf{Datasets}\\
         \midrule
\textbf{Logical} 
& bigbench repeat copy logic, mmmlu answer generation \\
\textbf{Causal}
& plausable result generation, anli r2 entailment, anli r3 entailment, cb entailment \\
\textbf{Commonsense}
& piqa answer generation, commongen sentence generation, sciq answer generation, openbookqa question answering\\
\textbf{Entailment}
& nli r2 entailment, anli r3 entailment, cb entailment, lue entailment classification 
\\
\textbf{Mathematics} & semeval closed vocabulary math, semeval geometric math, mmmlu formal logic\\
\textbf{Abductive} & tellmewhy \\
\textbf{Spatial} & babi t1 single supporting fact, piqa answer generation, toqa find location easy clean \\
\textbf{Analogical} & commongen sentence generation, bard analogical reasoning causation \\
\textbf{Argument} & argument stance classification, argument consequence classification \\
\textbf{Deductive} & rocstories correct answer generation \\
\bottomrule
    \end{tabular}}
    \caption{\footnotesize  \ourmodel{} benchmark consists of 20 datasets covering 10 different reasoning skills. The full list of the reasoning skills and datasets is in Table~\ref{tab:benchmark_full} in Appendix~\ref{sec:app_reasoning_benchmark}.}
    \label{tab:benchmark}
\end{table}

\paragraph{Summary of findings:} 
(\textit{i}) Different from \citet{gururangan2020don}, our experiments indicate that there is no strong correlation between high vocabulary overlap (between finetuning and evaluation datasets) and performance gain on reasoning evaluation datasets.  
This means that LLMs are not simply memorizing the training data during the finetuning stage; 
(\textit{ii}) Finetuning helps improve certain reasoning capabilities of LLMs (e.g. analogical and abductive) but not all of them (e.g. commonsense reasoning);
(\textit{iii}) Finetuning can cause overfitting towards data format, which makes it harder for LLMs to generalize to other prompt templates, while CoT-finetuning helps to mitigate this issue as it incorporates a variety of explanations.  



Though many of the aspects that we study have been discussed in prior analyses of LLMs~\citep{chung2022scaling,flan, cot, kojima2022large, cobbe2021training, sanh2021multitask}, prior work has not evaluated LLMs on different reasoning skills and how these skills can be improved. Overall, by evaluating reasoning skills with \ourmodel, we gain new insights on how models have or have not succeeded in generalizing beyond their \textit{training} experience.

To summarize our contributions, this paper presents a meticulously designed benchmark for assessing reasoning abilities. Furthermore, a thorough investigation of \textit{the role of finetuning} in the context of reasoning abilities, data memorization, and data format is conducted.


\section{Motivation and Our Benchmark}
\label{sec:reasoning_benchmark}

\paragraph{Motivation.}The analyses in \ourmodel~are inspired by a scientific question: To what extent do
LLMs learn generalizable reasoning abilities? This question motivates our focus on measuring LLMs' performance on tasks that require contextual understanding and perform multi-step operations, which are crucial to perform well on downstream tasks.

\begin{figure}[t]
\centering
\begin{boxedminipage}{0.9\columnwidth}
\vspace{-0.3cm}
\begin{equation*} 
    \scriptsize
    \begin{split}
        \hspace{0cm} & \text{\scriptsize \color{purple}{Definition:}\;  }\color{black}{\text{In this task, we ask you to write an implausible}}\\
        & \text{answer to a question that involves event duration, based on a given}\\
        & \text{sentence. Here, event duration is defined as the understanding of}\\
        & \text{how long events typically last. For example, “brushing teeth”,} \\
        & \text{usually takes a few minutes. Even though there exist multiple} \\
        & \text{wrong answers, we only need a single wrong answer.}\\
        & \text{\scriptsize \color{purple}{Example\;1-} } \\ 
        & \hspace{0.4cm} \text{\scriptsize \color{purple}{input:}\;} \color{black}\text{Sentence: Jack played basketball after school, after} \; \\
        & \hspace{0.4cm} \text{which he was very tired.}\\
        & \hspace{0.4cm} \text{Question: How long did Jack play basketball?}\\
        & \hspace{0.4cm} \text{\scriptsize \color{purple} output:\; } \color{black}\text{22 hours.} \; \\ 
        & \hspace{0.4cm} \text{\scriptsize\color{purple}explanation: } \;\color{black}\text{Typically we play basketball for a couple of}\\
        & \hspace{0.4cm} \text{hours. So any answer beyond that range is unlikely.} \\ 
    \end{split}
\end{equation*}
\end{boxedminipage}
\caption{\footnotesize An example from NIV2 \cite{niv2} that requires a deep understanding of the long task instruction and can be very challenging even for humans. 
}
\vspace{-3mm}
\label{tab:niv2_examples}
\end{figure}

\paragraph{Datasets Construction.} To construct the datasets of \ourmodel, we select datasets from NIV2 benchmark \cite{niv2} and perform the following operations:

\noindent{\textbf{(1) Omit extremely hard tasks.}} We design \ourmodel~so that it can be used to benchmark a variety of LLMs, from pre-trained, finetuned to instruction-tuned models. 
To select such tasks, we apply several heuristics: firstly, we manually omit tasks that heavily rely on instructions. Some tasks are hard to solve when only in-context examples (demonstrations) are provided (e.g., the example in Figure~\ref{tab:niv2_examples}). 
Secondly, we selected only those tasks that achieved a reasonable level of performance (empirically use ROUGE-L $>$ 5.0) when evaluated with a pre-trained model (we use the OPT-13B model).
Thirdly, we omit tasks on which humans fail to get decent performance given the ground truth labels from NIV2. For example, \textit{task963\_librispeech\_asr\_next\_word\_ prediction} \cite{weir2020cod3s} provides a prompt ``Joey's favourite food is \_\_\_'', with the ground truth answer ``sandwiches''. Without any context or background information, the answer can be any food thus it is extremely hard for humans to accurately predict ``sandwiches''.

\noindent{\textbf{(2) Remove tasks with long input context}}. The input sentence length of some tasks can be very long, and currently most LLMs are not designed for solving long text problems. We omit tasks with demonstration length longer than 2048 tokens.

\noindent{\textbf{(3) Fix ground truth labels.}} For each reasoning task, NIV2 provides the reasoning skills required to solve the task, e.g. \textit{task102\_commongen\_data\_to\_text} requires relational, analogical and commonsense reasoning. However, we found that some tasks have been labeled with incorrect reasoning skills. For example, \textit{task393\_plausible\_result\_generation} provides a sentence and asks LLMs to complete the sentence. The labels given by NIV2 are causal reasoning and textual entailment, but in fact this task can hardly examine an entailment skill. Accordingly, we manually fix reasoning skill labels. In addition, we only keep the predominant skill. For example, many tasks need more or less commonsense knowledge, therefore we select the related tasks that only heavily rely on commonsense knowledge to assess commonsense reasoning.

\paragraph{Benchmark.}
After the above steps, we select tasks that represent a variety of reasoning skills and construct \ourmodel~reasoning benchmark, where Table~\ref{tab:benchmark} shows details about our benchmark.

\section{Experiment Setup}
\label{sec:fnetuning}
\subsection{Models}
\begin{figure}[t]
    \centering
    \vspace{-0.2cm}
    \includegraphics[width=\linewidth]{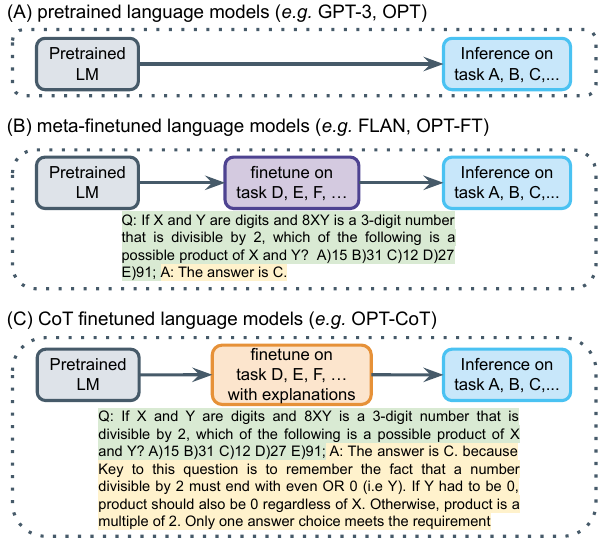}
    \caption{\footnotesize  We compare three types of models: (A) directly apply pretrained LLMs on reasoning tasks; (B) finetune LLMs on a set of tasks; (C) finetune LLMs on tasks with explanations (CoT-finetuning). Finetuning data contains \colorbox{lightgreen}{source} and \colorbox{lightyellow}{target} parts, and the language modeling loss only applied to the \colorbox{lightyellow}{target} part.}
    \vspace{-3mm}
    \label{fig:baselines}
\end{figure}

To perform a controlled comparison across training and prompting methods, we focus on three different models: pre-trained, meta-finetuned, and rationale-based meta-finetuned (CoT-finetuned) models. 
For pre-trained models, we use OPT \cite{opt}, a suite of decoder-only pre-trained transformers which are reported to yield comparable performance to GPT-3 \cite{brown2020language}. We benchmark with OPT models of two scales: 1.3B and 13B. For finetuned models (OPT-FT), we finetune OPT models on datasets without explanations. For CoT-finetuned models (OPT-CoT), we finetune OPT models on data with rationales (explanations). 

We train all models in Pytorch \cite{paszke2017automatic} using OPT-IML \cite{iyer2022opt} codebase\footnote{\scriptsize{\url{https://github.com/facebookresearch/metaseq/tree/main/projects/OPT-IML}}}. We initialize model hyper-parameters for each model scale following OPT \cite{opt}. We pack our training examples into sequences of length $2048$, left-truncating examples that overflow. We use AdamW \citep{loshchilov2018fixing} with 32-bit state with $(\beta_1, \beta_2) = (0.9, 0.95)$, linearly warming up the learning rate for $6\%$ steps to the maximum, followed by linearly decaying it to 0. For all 1.3B models, we use batch size of 128, and for 13B models, we use batch size of 256.

\subsection{Finetuning Data}
\label{finetuningdetails}
Our finetuning corpus is comprised of 10 datasets: ProofWriter \cite{tafjord2020proofwriter}, StrategyQA \cite{geva2021did}, ECQA \cite{aggarwal-etal-2021-explanations}, CoQA \cite{reddy-etal-2019-coqa}, GSM8K \cite{cobbe2021training}, AQUA-RAT \cite{ling-etal-2017-program}, ESNLI \cite{camburu2018snli}, MATH \cite{hendrycks2021measuring}, CoS-E \cite{rajani2019explain}, WinoWhy \cite{zhang-etal-2020-winowhy}. These 10 finetuning datasets collectively contain 6 different reasoning skills: logical reasoning, causal reasoning, commensense reasoning, textual entailment, mathematics, abductive reasoning. In addition, these 10 datasets all come with instructions, demonstration examples and explanations. This enables fair comparison of OPT-FT and OPT-CoT models. More details about finetuning corpus can be found in Table~\ref{tab:app_training_corpus} in Section~\ref{sec:app_training_corpus}. More details about development data selection can be found in the  Appendix.~\ref{sec:app_devset}.

\subsection{Evaluation}
\label{sec:evaluation}



\paragraph{Templates} 

Following \cite{wei2021finetuned}, to control for the effect of variable prompt templates, we adopt different templates (\textbf{T}) during inference stage in our experiments:

\noindent\textbf{T1}: instruction + demonstration examples with explanations + "let's think step by step";

\noindent\textbf{T2}: instruction + "Please give a short explanation after the answer" + demonstration examples with explanations + "let's think step by step"

\noindent\textbf{T3}: instruction + "Please give a short explanation after the answer" + demonstration examples with explanations

\noindent\textbf{T4}: "Please give a short explanation after the answer" + demonstration examples with explanations + "Let's think step by step"

\noindent\textbf{T5}: instructions + demonstrations


For each dataset, we 
report the average and max score among these five templates. The final aggregated results (including aggregated average score and aggregated max score) are reported by further averaging across all datasets. Unless specified otherwise, the default score refers to the aggregated max score among five templates.

\paragraph{Evaluation metrics.} 
Since our benchmark contains both classification and generation tasks, we cannot use classification accuracy to evaluate all the tasks. Following FLAN \cite{wei2021finetuned}, we append classification choices at the end of prompts and ask models to generate answers. Thus, classification tasks can be treated as a special case of generation tasks. 
Accordingly, we use ROUGE-L \cite{lin2004rouge} to measure the performance of both classification and generation tasks and report the aggregated score. Similar to \citet{chung2022scaling}, we also use \textit{exact-match} score which is more suitable for tasks with short answers. Additionally, we compute \textit{relaxed-match} score which is a relaxed version of exact-match. Specifically, we normalize ground truth answers and predictions to have all text in lower case and remove punctuation and extra white spaces.


\section{Analysis}
\label{sec:analysis}
\subsection{Does finetuning help?}


\begin{figure*}[t]
\centering
\begin{minipage}{.67\textwidth}
  \centering
  \includegraphics[width=\linewidth]{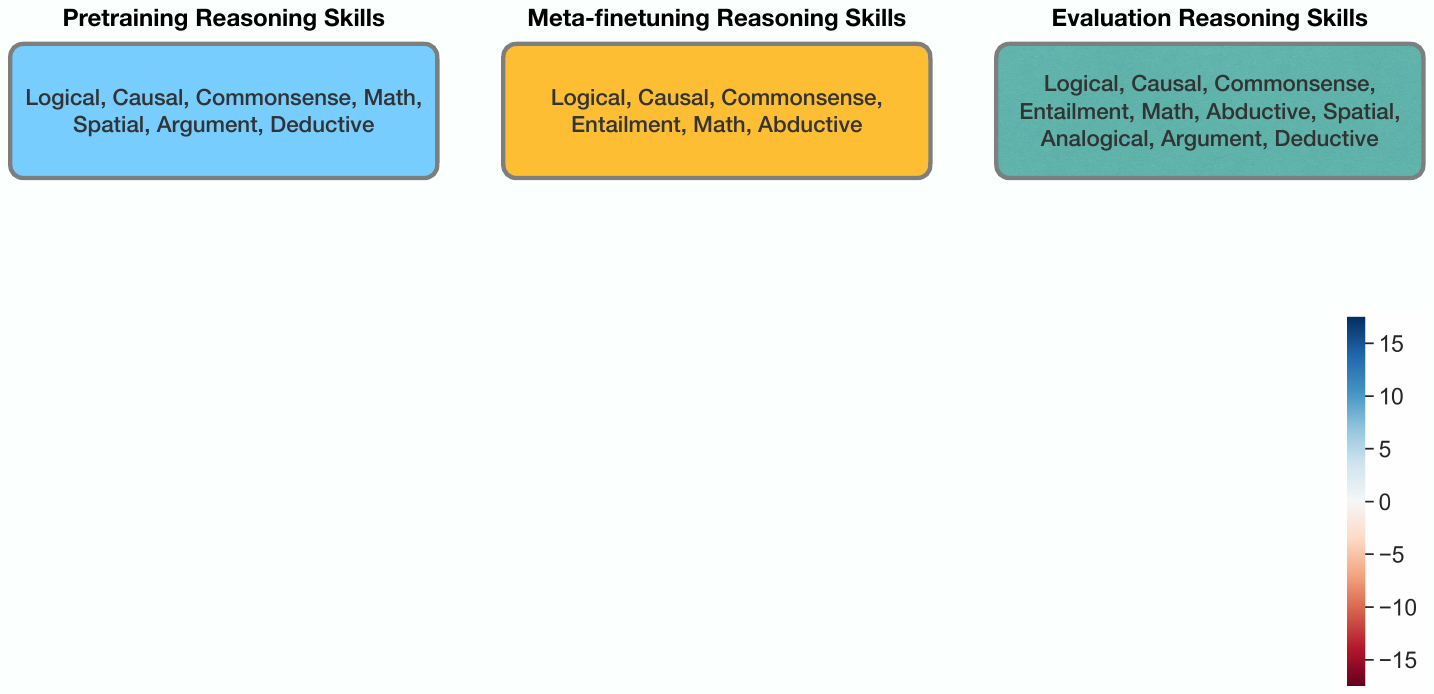}
  \captionof{figure}{\footnotesize Performance of pre-trained LM (OPT), finetuned LM (OPT-FT) and CoT-finetuned LM (OPT-CoT) on \ourmodel~reasoning benchmark. Left charts show aggregated \textbf{max} scores while right are \textbf{average} scores across 5 templates. Scores are averaged across 20 tasks.}
  \label{fig:overall_performance}
\end{minipage}%
\hspace{0.5cm}
\begin{minipage}{.29\textwidth}
  \centering
  \includegraphics[width=\linewidth]{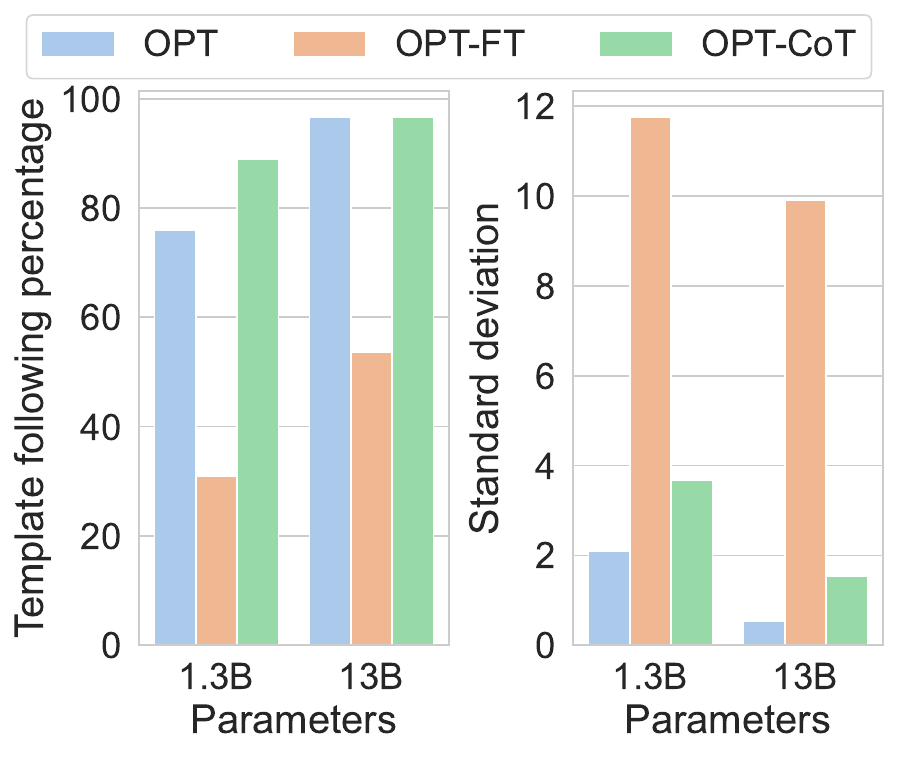}
  \captionof{figure}{\footnotesize Analyzing the robustness of models in following the templates. \textbf{Left}: template following percentage by each model; \textbf{Right}: standard deviation of template following percentage.}
  \label{fig:follow_rate}
  \vspace{-6mm}
\end{minipage}
\end{figure*}

Figure~\ref{fig:overall_performance} demonstrates the performance averaged across all evaluation tasks in our benchmark. 
Rationale-based finetuning (OPT-CoT) has been shown to improve the performance of the 1.3B model by 3.89\% in terms of the aggregated max ROUGE-L score and 3.83\% in terms of the aggregated max exact-match score. As for 13B model, OPT-CoT gains the improvement by 15.22\% in regard of aggregated max ROUGE-L score, 12.64\% in regard of aggregated max exact-match score. However, finetuning (OPT-FT) sometimes yields worse results than the vanilla pre-trained model. 




\subsection{What does LLMs learn during finetuning?}

We find that CoT-finetuning improves performance on reasoning tasks in general. However,
what exactly does the LLMs learn during the finetuning stage is still under explored. Thus, we study the role of finetuning from three perspectives: data memorization, reasoning skill transfer, and prompt template memorization.


\subsubsection{Data Memorization}
\label{sec:vocabulary_overlaps}

\begin{figure}[t!]
\centering
\scalebox{1.0}{
    \includegraphics[width=\linewidth]{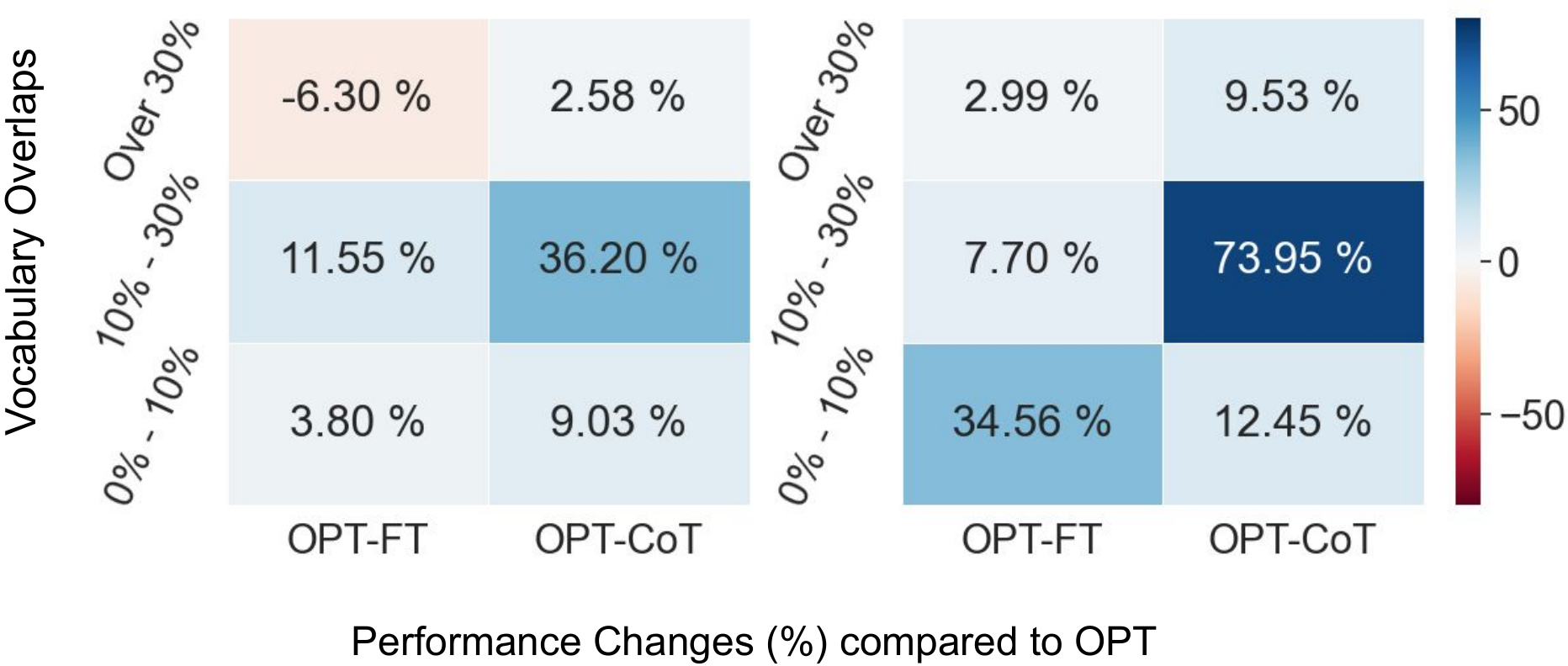}
}
    \caption{\footnotesize Correlation between \textbf{vocabulary overlap} and \textbf{performance improvement} using 13B parameter models. The \textbf{left} chart shows ROUGE-L while the \textbf{right} shows relaxed-match score.}
    \vspace{-3mm}
    \label{fig:vocabulary_overlaps}
\end{figure}

\begin{figure*}
\centering
\begin{minipage}{.27\textwidth}
  \centering
  \includegraphics[width=\linewidth]{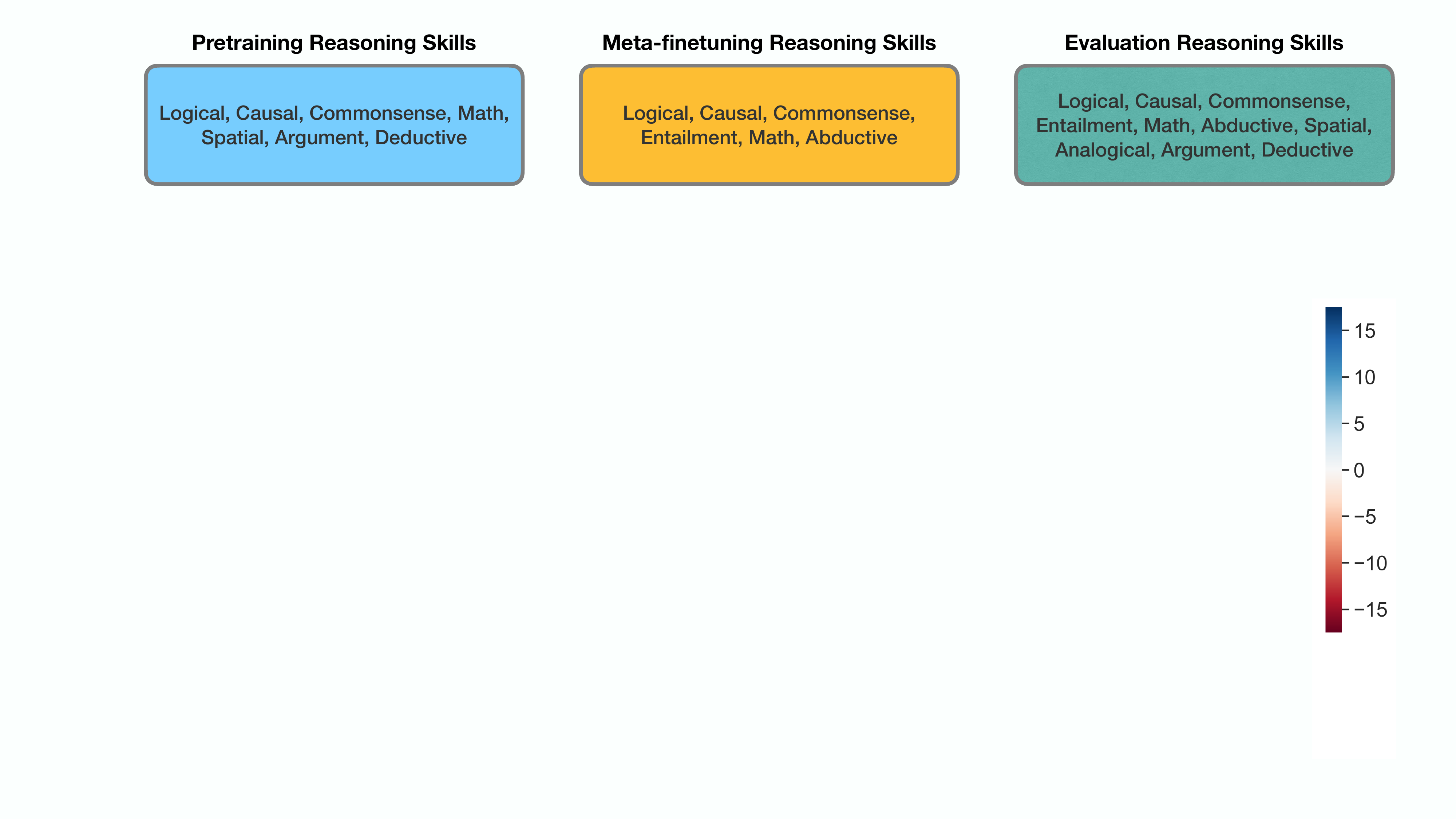}
  \captionof{figure}{\footnotesize Reasoning skills learned during pretraining and meta-finetuning stages, as well as tested through \ourmodel~.}
  \label{fig: teaser}
\end{minipage}%
\hspace{0.5cm}
\begin{minipage}{.69\textwidth}
  \centering
  \includegraphics[width=\linewidth]{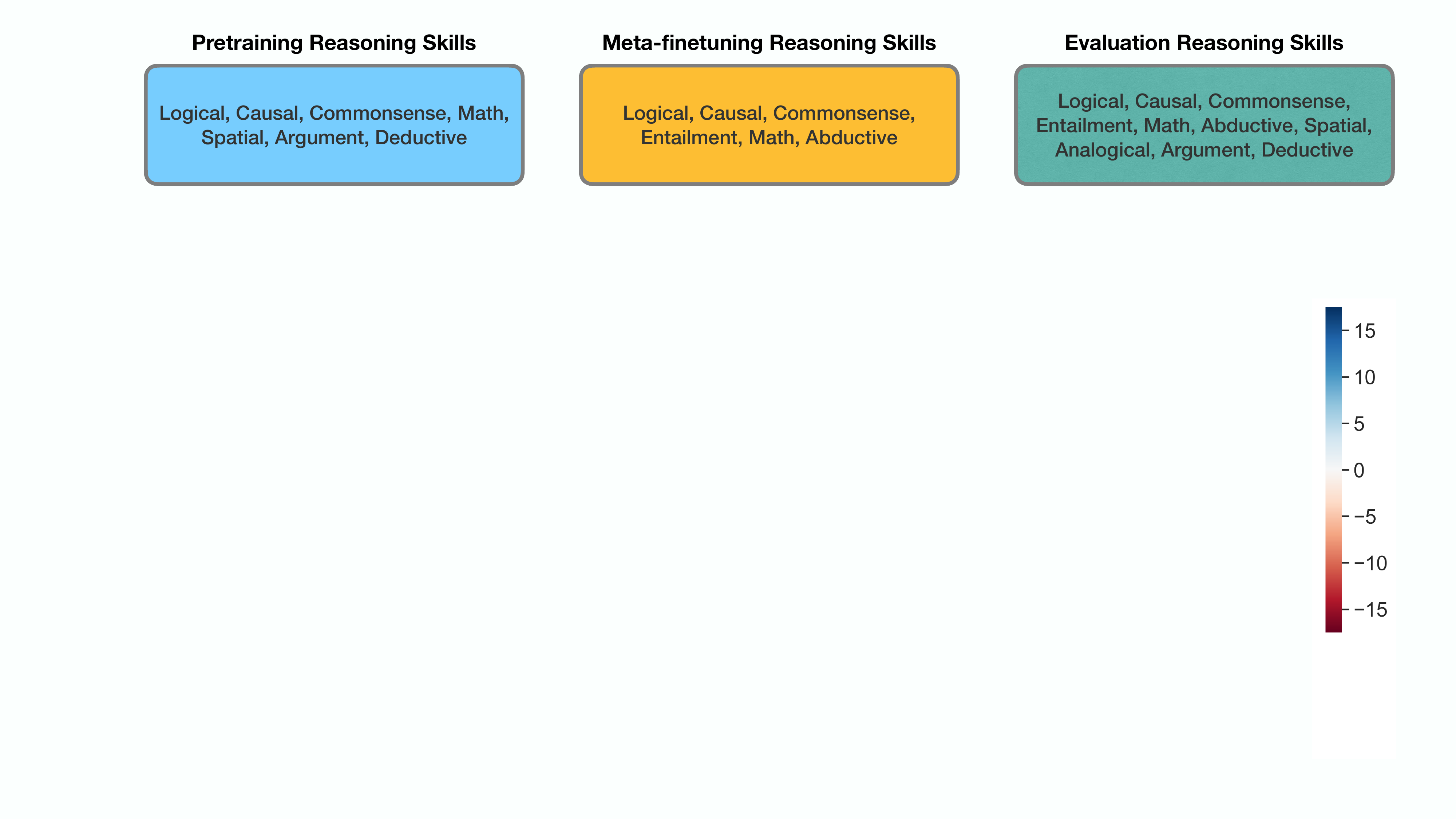}
  \vspace{-8mm}
  \captionof{figure}{\footnotesize The ROUGE-L scores illustrating the difference between OPT-FT and OPT, as well as OPT-CoT and OPT models within each reasoning skill. Left: skills split by pretraining data; Right: skills split by meta-finetuning data.}
  \label{fig:analyze_reasoning_skills}
\end{minipage}
\vspace{-3mm}
\end{figure*}

\citet{gururangan2020don} finds that the performance gain is larger when the finetuning dataset is more dissimilar to the pre-training dataset. However, their conclusion is made by a single-task finetuning. They evaluate their model on the same dataset that was used for finetuning. A more thorough evaluation dictates that finetuned models \cite{wei2021finetuned,chung2022scaling} be evaluated on held-out datasets. As such, in Figure~\ref{fig:baselines} in blocks (B) and (C) we show two potential ways of finetuning and inference as illustrated here in our paper. 


To confirm that the improvement in finetuning performance is due to the increased amount of data seen during the finetuning stage, we measure the dissimilarity between the training data used in finetuning and evaluation, respectively.
If higher similarity leads to better performance, it may indicate that the improvements of finetuned LLMs are due to seeing more similar data during the finetuning stage. Following \cite{gururangan2020don}, we use unigram vocabulary overlap to measure the data similarity.    
More specifically, we divide our tasks into three categories: The first category has 10 datasets which consists of up to 10\% overlap between the finetuning data and evaluation data. The second category comprises 3 datasets with an overlap between 10\% and 30\%. The third category has 7 datasets with an overlap over 30\%. 
Details can be found in Table~\ref{tab:full_vocabulary_overlaps} in appendix~\ref{sec:app_vocabulary_overlaps}.

We measure the performance improvements of OPT-FT and OPT-CoT compared against the pretrained OPT model.
We present both ROUGE-L score (top) and relaxed-match score (down) in Figure~\ref{fig:vocabulary_overlaps}.
The results indicate that there is no strong correlation between the vocabulary overlap between fineuning and evaluation datasets and the performance of the model (neither a higher nor a lower vocabulary overlap always translate to a performance improvement). OPT-CoT achieves the best ROUGE-L and relaxed-match scores both in settings when there is a medium (10\%-30\%) level of vocabulary overlap. We don't observe a consistent pattern on OPT-FT models either. Overall, 
for these challenging tasks, seeing similar data during finetuning stage does not guarantee performance improvement. 



\subsubsection{Reasoning Skill Transfer}
\label{sec:reasoning_skills}



Table~\ref{fig: teaser} illustrates the reasoning skills present in each stage. 7 skills can be learned from pretraining data.
Appendix.~\ref{sec:app_pretrain_data} shows more details about pretraining data. 6 skills can be learned from finetuning data (Table~\ref{tab:app_training_corpus}). Using \ourmodel~ we measure a total of 10 reasoning skills in model evaluation.

The average ROUGE-L scores are calculated for each reasoning skill on 6 models (1.3B OPT, 1.3B OPT-FT, 1.3B OPT-CoT, 13B OPT, 13B OPT-FT, 13B OPT-CoT). Figure~\ref{fig:analyze_reasoning_skills} shows the difference between OPT-FT and OPT, and the difference between OPT-CoT and OPT models' performance. For example, OPT-FT 1.3B model yields on average 3.5 less ROUGE-L points than OPT 1.3B model on the tasks of logical reasoning.


Figure~\ref{fig:analyze_reasoning_skills} contains 4 sub-figures, showing reasoning skills transfer results: 
(\textit{i}) The upper left sub-figure shows 7 skills that are acquired during the pretraining stage (OPT pretraining data), and how much improvement can be obtained through meta-finetuning (OPT-FT and OPT-CoT); 
(\textit{ii}) The bottom left sub-figure illustrates that these 3 skills are harder to acquire during the pre-training stage, and the amount of improvement that can be obtained through meta-finetuning;
(\textit{iii}) The upper right sub-figure illustrates that such 7 skills are acquired during the meta-finetuning stage through finetuning datasets (Table~\ref{tab:app_training_corpus}). Do these skills show improvement measured by evaluation benchmark? 
(\textit{iv}) The bottom right sub-figure studies the reasoning skills that were not learned in the finetuning stage, can these skills be improved through meta-finetuning? We study the answers to these questions below.

From figure (\textit{ii}) We observe that all four of the LLMs demonstrate enhanced reasoning capabilities on textual entailment, abductive reasoning, and analogical reasoning tasks. These abilities are not readily acquired during the pretraining stage, as the pretraining data consists only of plain text. On the other hand, skills such as commonsense reasoning or spatial reasoning can be gained during the pretraining stage, while the benefits of further finetuning are not as pronounced. Additionally, \citet{gururangan2020don} concluded that the more dissimilar the domain between pretraining and finetuning are, the higher the potential for finetuning to yield gains. We see the same trend but the domain in \citet{gururangan2020don} is defined by the vocabulary overlaps, while we define the domains by reasoning skills. From figure (\textit{iii}) we can see that the reasoning skills gained during the meta-finetuning stage may not necessarily transfer to the improvement of the same skills on the evaluation datasets.


We also observe that finetuning with OPT-CoT enables the model to acquire a wider range of reasoning skills, resulting in stronger performance on logical and causal reasoning tasks, in addition to skills that consistently improve across all finetuned models.


\subsubsection{Data Format Memorization}
\label{sec:data_format}

\begin{figure}[t!]
\centering
\scalebox{0.70}{
    \includegraphics[width=0.45\textwidth]
    {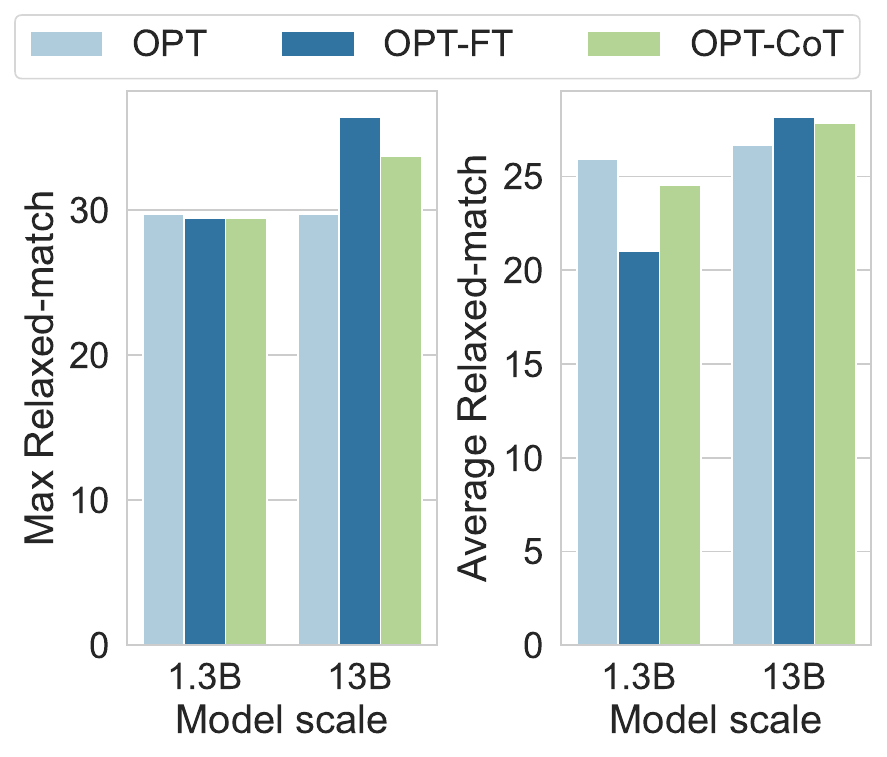}}
    \vspace{-3mm}
    \caption{\footnotesize Comparing pretraining and finetuning models with relaxed match score. \textbf{Left}: aggregated best (max) performance across 5 Templates; \textbf{Right}: aggregated average performance across 5 Templates.}
    \vspace{-3mm}
    \label{fig:relaxed_match}
\end{figure}

We investigate whether finetuning can simply memorize the template representation of the training data, and the effect of data format on the robustness of the models.

\paragraph{Evaluation with relaxed-match score.}
We compare two metrics: exact-match and relaxed-match. From Figure~\ref{fig:overall_performance}, we observe that OPT-FT is worse than OPT when exact-match is used as the metric. However, when relaxed-match is used, OPT-FT outperforms OPT as shown in Figure~\ref{fig:relaxed_match}. Relaxed-match score ignores punctuation, articles and extra whitespace. This suggests that if we decouple performance from format adherence, OPT-FT performs better than OPT. In other words, finetuning is helpful but it can make the output more noisy. This explains the reason for the performance drop when exact-match is used as the metric.


\paragraph{Template following percentage.} We check whether the model can follow the template of the demonstrations. For example, if a demonstration uses "the answer is xxx because yyy", then we check what percentage of instances can follow the exact same template as the demonstration.  Figure~\ref{fig:follow_rate} (left) shows the average template following percentage for each model. 
Both OPT and OPT-CoT consistently show that they can follow demonstrations' even though OPT is not pre-trained on rationales. Compared to 1.3B models, 
larger models demonstrate a greater overall ability to follow the template of the demonstrations.
Compared to OPT and OPT-CoT, OPT-FT lacks the ability to follow diverse templates. This is because the OPT-FT training process does not contain any rationale data. 
Finetuning causes the model to become more biased towards a particular template representation, while its ability to adapt to other templates becomes impaired. It is worth noting that despite being trained on rationales, the OPT-CoT model performs well when evaluated using non-CoT templates.

    

\paragraph{Robustness} 
To assess the robustness of each model to various templates, we compute the standard deviation of ROUGE-L scores for each model across five different templates. As we can see from Figure~\ref{fig:follow_rate} (right), OPT is robust to different templates, while OPT-FT has difficulties adapting to changing templates. In general, finetuning (both OPT-FT and OPT-CoT) adversely affects the robustness of the model and makes the model biased towards a specific data format, however, OPT-CoT is better than general finetuning (OPT-FT).

\paragraph{Reasoning chain quality.}
Following \cite{golovneva2022roscoe} we evaluate reasoning abilities of the models using \texttt{ROSCOE} scoring suite (Table~\ref{tab:roscoe-sum}). Looking at each score in detail (Appendix~\ref{sec:app_chains_eval}), we found that overall across templates OPT-FT models produce shorter, less informative chains, while OPT baseline models produce long chains with high amount of self-repetitions. 13B OPT-CoT chains showed best quality despite some self-consistency and grammar issues. When comparing prompt templates, models prompted with Template 5 produce short chains, often without reasoning at all, even if they were fine-tuned on reasoning chains (OPT-CoT), suggesting overfitting to the prompt template.
\begin{table}[t!]
\begin{center}
\setlength\tabcolsep{3.2pt}
\scalebox{0.7}
    {
 \begin{tabular}{l|ccc|ccc}
    \toprule
    & \multicolumn{3}{c}{1.3B} & \multicolumn{3}{c}{13B}\\
  Metrics & OPT & OPT-FT & OPT-CoT & OPT & OPT-FT & OPT-CoT \\
\hline
\texttt{ROSCOE-SA} & 0.936 & 0.921 & \underline{0.938} & 0.936 & 0.923 & \textbf{0.940} \\
\texttt{ROSCOE-SS} & \underline{0.925} & 0.923 & 0.920 & \textbf{0.926} & 0.916 & \underline{0.925} \\
\texttt{ROSCOE-LI} & 0.848 & \textbf{0.953} & 0.875 & 0.863 & \underline{0.944} & 0.890 \\
\texttt{ROSCOE-LS} & \underline{0.725} & \textbf{0.744} & 0.666 & 0.688 & 0.705 & 0.640 \\
    \bottomrule
    \end{tabular}%
    }
	\end{center}
 \vspace{-3mm}
 \caption{\footnotesize Summary of the \texttt{ROSCOE} evaluation results averaged across templates. Each metric is bounded within $[0, 1]$, where $1$ indicates the perfect score and $0$ corresponds to failure. In each row, values corresponding to the best-performing model are \textbf{bolded}, second best are \underline{underscored}.}
 \vspace{-3mm}
\label{tab:roscoe-sum}
\end{table}

In summary, models learn the data format representation and templates during finetuning stage. However, finetuned models contain bias towards the data formats and template it has seen, which potentially reduces the robustness of the model to more generalized settings. When comparing robustness, OPT-CoT is better than OPT-FT, but it is still not as robust as the pre-trained model. 


\section{Related Work}
\label{sec:related_work}
\label{main-relatedwork}
\paragraph{LLMs that Reason.} 
To improve LLMs' reasoning abilities, \citet{kojima2022large} shows that LLMs can be decent zero-shot reasoners by simply appending ``Let’s think step by step'' to the prompt. ~\citet{cot} adds a series of intermediate reasoning steps to improve LLMs' reasoning abilities. ~\citet{wang2022rationale} further proposes to expand prompts to include rationales in each few-shot example. \citet{fu2022complexity} discovers that prompting with higher reasoning complexity achieves substantial gains on math word tasks. To tackle problems harder than demonstration examples, ~\citet{zhou2022least} first reduces a complex problem into a list of subproblems and solve subproblems sequentially. Another line of research is to improve the naive decoding strategy, \citet{wang2022self} introduces a self-consistency strategy which selects the most consistent answer among a set of reasoning paths.

\paragraph{Existing Reasoning Benchmarks.}

Many benchmarks are used for evaluating language models' performance, such as BIG-Bench \cite{srivastava2022beyond}, Natural Instruction V2 (NIV2) \cite{niv2}, MMLU \cite{hendrycks2020measuring}. Although they contain some reasoning tasks, none of them are specifically designed to test models' reasoning skills. For example, NIV2 contains 172 datasets and a total of 1554 tasks, including some reasoning tasks. It has several issues which make it inappropriate to be directly used as a reasoning benchmark:  (\textit{1}) it is designed for instruction-tuned models and some tasks might be unsuitable for evaluating pretrained models or non-instruction finetuned models, as shown in Figure~\ref{tab:niv2_examples}; 
(\textit{2}) reasoning skills have been divided into 27 categories while some of them have large overlaps, e.g. numerical reasoning, quantitative reasoning, reasoning on numbers; 
(\textit{3}) some reasoning labels are wrongly labeled, e.g. \textit{task393\_plausible\_result\_generation} gives textual entailment label but this task can hardly examine the entailment skill.

The Curriculum benchmark \cite{chen2022curriculum} is designed for probing LLMs' reasoning abilities and covers $8$ different reasoning skills. However, this work only focuses on classification tasks and it converts all examples into the Natural Language Inference (NLI) format to fit into a unified framework. We argue that the forced conversion of all datasets into the NLI format does not align with human natural conversational style. We observed that even davinci-003 fails at some simple tasks due to their forced conversion, e.g. examples in Table~\ref{tab:reasoning_examples}. More discussion and results are shown in the Appendix~\ref{sec:app_curriculum}.



\paragraph{Finetuning LLMs.}
LLMs meta-finetuned on a range of NLP tasks have shown improved performance on held-out downstream tasks such as FLAN \cite{wei2021finetuned}, T0 \cite{sanh2021multitask}, Tk-Instruct \cite{niv2} and Instruct-GPT \cite{ouyang2022training}. Following this approach, we finetune OPT models and name this type of models as OPT-FT ((B) in Figure~\ref{fig:baselines}). \citet{chung2022scaling} further adds chain-of-thought data at finetuning stage and shows significant improvements. We also study this type of models and name them as OPT-CoT ((C) in Figure~\ref{fig:baselines}). However, from previous research it still remains unclear whether the improvement comes from simply adding more training data or finetuning on rationales actually helps. We conduct rigorous evaluations to address this question.


\section{Conclusion}
\label{sec:conclusion}
We introduce \ourmodel, a carefully curated benchmark for evaluating reasoning abilities of LLMs. 
It comprises over 20 datasets and covers 10 different reasoning skills. Using this benchmark, we further investigate the impact of finetuning on these complex tasks. Our experiments reveal that 
LLMs do not simply memorize training data, but are capable of learning various reasoning skills, 
such as textual entailment, abductive reasoning and analogical reasoning. While we found that finetuning generally leads to improved performance, we also discovered some negative effects. LLMs tend to memorize the data template representation and templates seen during finetuning, thus reducing the robustness of the model to generalized settings. CoT-finetuning (OPT-CoT) can alleviate this issue to some extent, but it is still less robust compared to the vanilla pre-trained model.

\section*{Limitations}


\ourmodel~ aims to encompass a wide range of reasoning skills, but some reasoning skills are missing, specifically in regards to symbolic reasoning (last letter concatenation task and coin flip \cite{cot}) and compositionality reasoning (SCAN \cite{lake2018generalization}, COGS \cite{kim2020cogs} and CFQ \cite{keysers2019measuring}). These reasoning skills should be included in future work. 

In terms of computing power, we have experimented with models that were accessible to us. We acknowledge that there are larger models that we were not able to train due to the limitations of our computational budget.


During our analysis, we discovered that some datasets contain noise, where even human experts are unable to provide accurate answers for certain instances. While it is important to address this issue, it is a time-consuming process to carefully review and clean each instance in the dataset. We plan to address this in future work.



\section*{Ethics Statement}

Large language models (LLMs), due to potential bias in the training data, can be prone to generate toxic and unwanted content \cite{weidinger2021ethical}. However, in this paper, we are focused on reasoning tasks where the model is prompted to explain its decisions, because of which our model falls under contained generation. By providing clear prompts and constraints, we believe that this might help guide the model's output towards specific, desired outcomes and reduce the likelihood of generating unwanted or harmful content, as opposed to open ended text generation tasks.


\bibliography{custom}
\bibliographystyle{acl_natbib}

\newpage

\appendix

\section{More Details about Data Usage}
\label{sec:app_detailed_data}

\subsection{Reasoning Benchmark}
\label{sec:app_reasoning_benchmark}

Table~\ref{tab:benchmark_full} shows detailed reasoning benchmark. 

\begin{table*}
\centering
\scalebox{0.85}{
\begin{tabular}{l| c |c}
    \toprule
        \makecell{\textbf{Reasoning} \\ \textbf{Skills}} & \textbf{Task ID} & \textbf{Datasets}  \\
        \midrule
         \makecell{Logical \\Reasoning}  & \makecell{62\\697} & \makecell{bigbench repeat copy logic \cite{srivastava2022beyond}\\  mmmlu answer generation formal logic \cite{hendryckstest2021}} \\
         \midrule
          \makecell{Causal \\Reasoning} & \makecell{393\\1386\\1387\\1388}&\makecell{ plausible result generation \cite{weir2020cod3s} \\ anli r2 entailment \cite{williams-etal-2020-anlizing} \\ anli r3 entailment \cite{williams-etal-2020-anlizing}\\ cb entailment \cite{wang2019superglue} } \\
          \midrule
          \makecell{Commonsense \\Reasoning} &\makecell{80\\102\\591\\1286}& \makecell{piqa answer generation \cite{Bisk2020} \\ commongen sentence generation \cite{lin-etal-2020-commongen} \\ sciq answer generation \cite{welbl2017crowdsourcing} \\ openbookqa question answering \cite{mihaylov2018can}}\\
          \midrule
          \makecell{Texual \\Entailment}&\makecell{1386\\1387\\1388\\1344} & \makecell{anli r2 entailment \cite{williams-etal-2020-anlizing} \\ anli r3 entailment \cite{williams-etal-2020-anlizing} \\ cb entailment \cite{wang2019superglue} \\ glue entailment classification \cite{wang2019glue}} \\
          \midrule 
          Mathematics & \makecell{104\\119\\697}&\makecell{semeval closed vocabulary math answer generation \cite{hopkins2019semeval}\\ semeval geometric math answer generation \cite{hopkins2019semeval}\\ mmmlu answer generation formal logic \cite{hendryckstest2021}}\\
          \midrule
          \makecell{Abductive \\Reasoning} &332& tellmewhy answer generation \cite{lal-etal-2021-tellmewhy} \\
          \midrule
          \makecell{Spatial \\Reasoning} & \makecell{83\\80\\151}&\makecell{babi t1 single supporting fact answer generation \cite{weston2015towards}\\ piqa answer generation \cite{Bisk2020} \\ tomqa find location easy clean \cite{nematzadeh2018evaluating}}  \\
          \midrule
          \makecell{Analogical \\Reasoning} &\makecell{102\\ 1152}& \makecell{commongen sentence generation \cite{lin-etal-2020-commongen} \\ bard analogical reasoning causation \cite{fulda2017harvesting}} \\
          \midrule
          \makecell{Argument \\Reasoning} &\makecell{513\\514}& \makecell{argument stance classification \cite{kobbe2020unsupervised} \\argument consequence classification \cite{kobbe2020unsupervised}} \\
          \midrule
          \makecell{Deductive \\Reasoning}&216 & rocstories correct answer generation \cite{mostafazadeh2016corpus} \\
         \bottomrule
    \end{tabular}}
    \caption{\footnotesize Details about \ourmodel~ benchmark.}\label{tab:benchmark_full}
\end{table*}

\subsection{Training Corpus (cont. from $\S$\ref{finetuningdetails})}
\label{sec:app_training_corpus}
We used 10 datasets for finetuning, which contain 6 different reasoning skills. 

\begin{table*}[ht]
    \centering
    \scalebox{0.8}{
    \begin{tabular}{c|c |c |c |c}
    \toprule
         Datasets & Train Size & Val Size & Test Size&  Reasoning Skills \\
         \midrule
         ProofWriter & 69,810 & 10,190 & 20,030 & Logical Reasoning, Causal Reasoning \\
         StrategyQA & 2,290 &-&490& Commonsense Reasoning \\
        ECQA & 7,598 &1,090 &2,194&Commonsense Reasoning \\
        CoQA & 10,8647 &7,983&-& Textual Entailment \\
        GSM8K & 7,473 & -&1,319&Mathematics \\
        AQUA-RAT & 97,467 &254 &254&Mathematics \\
        ESNLI & 549,367 & 9,842& 9,824&Commonsense Reasoning, Logical Reasoning, Textual Entailment \\
        MATH & 7,500 &-&5,000& Mathematics \\
        CoS-E & 9,741 & 1,221&-&Commonsense Reasoning \\
        WinoWhy & 273 & -&-&Abductive Reasoning, Commonsense Reasoning \\
         \bottomrule
    \end{tabular}}
    \caption{\footnotesize Training corpus for meta-finetuning OPT-FT and OPT-CoT. (Cont. from $\S$~\ref{finetuningdetails})}
    \label{tab:app_training_corpus}
\end{table*}

\subsection{Development Data Details}
\label{sec:app_devset}
Our finetuning models are tuned on pretrained LLMs on the finetuning corpus with the goal of improving the performance of unseen tasks. For example, blocks (B) and (C) in Figure~\ref{fig:baselines} are showing models that are finetuned on tasks B,C,D and the goal is to achieve good results on task A. 

Checkpoint selection can determine the final performance of the LLMs to a very large extent. There are several ways to select checkpoints: (\textit{i}) select checkpoint of the last iteration; 
(\textit{ii}) select checkpoint based on perplexity or loss from validation datasets of finetuning corpus (validation datasets of task B, C, D); 
(\textit{iii}) select checkpoint based on perplexity or loss from validation datasets of evaluation corpus (validation datasets of task A); 

In order to achieve a better performance on evaluation corpus, a common approach is to use methods like (\textit{iii}) to select a checkpoint. However, we would like to prevent LLMs overfiting to the distribution of our final evaluation corpus. We initially used the method (\textit{ii}) but found that it did't work well. However, this resulted in a distribution mismatch issue. We speculate this to the fact that some tasks in our finetuning corpus do not have a validation set. We thus select 3 tasks from NIV2 benchmark and compile a development set that does not have any overlaps with our finetuning data or evaluation data. 
There are 3 datasets used as our development set for checkpoint selection:
task 247 dream answer generation \cite{sundream2018}, task 118 semeval and task 10 open vocabulary mathematical answer generation \cite{hopkins2019semeval} and anli r1 entailment \cite{williams-etal-2020-anlizing}

\begin{table*}[ht]
    \centering
    \scalebox{0.8}{
    \begin{tabular}{c|c | c}
         \toprule
         \textbf{Task ID} & \textbf{Datasets} & \textbf{Reasoning Skills} \\
         \midrule
          247 & dream answer generation \cite{sundream2018} & \makecell{Logical Reasoning\\ Commonsense Reasoning}\\
          \midrule
          118 & \makecell{semeval open vocabulary mathematical\\ answer generation \cite{hopkins2019semeval}} & \makecell{Commonsense Reasoning\\ Mathematics}\\
          \midrule
          1385 & anli r1 entailment \cite{williams-etal-2020-anlizing} & \makecell{Textual Entailment\\Commonsense Reasoning \\Causal Reasoning} \\
          \bottomrule
    \end{tabular}}
    \caption{\footnotesize Dev set for checkpoint selection}
    \label{tab:app_devset}
\end{table*}

\subsection{Pretraining Data Analysis}
\label{sec:app_pretrain_data}


The pre-training corpus of OPT model \cite{opt} contains a concatenation of datasets used in RoBERTa \cite{liu2019roberta}, the Pile \cite{gao2020pile}, and PushShift.io Reddit \cite{baumgartner2020pushshift,roller2020recipes}.

\paragraph{RoBERTa}
Three datasets in RoBERTa \cite{liu2019roberta} are used as pretraining corpus: BookCorpus \cite{zhu2015aligning}, Stories \cite{trinh2018simple}, and CCNews \cite{liu2019roberta}. Deductive reasoning skill and spatial reasoning skill can be learned from stories dataset. Logical reasoning skill can be learned from these three datasets.

\paragraph{Pile}
A subset of the Pile \cite{gao2020pile} are used as pre-training corpus, including CommonCrawl, DM Mathematics, Project Gutenberg, HackerNews, OpenSubtitles, OpenWebText2, USPTO, and Wikipedia. Mathematics reasoning skill can be learned from DM Mathematics dataset. Causal Reasoning can be learned widely from OpenWebText2. Commensense reasoning skill can be learned from Wikipedia. 

\paragraph{PushShift.io Reddit}
The longest chain of comments in each thread are extracted from PushShift.io Reddit \cite{baumgartner2020pushshift}. Argument reasoning skill can be learned from this dataset.

\subsection{Vocabulary Overlaps (Cont. from $\S$~\ref{sec:vocabulary_overlaps})}
\label{sec:app_vocabulary_overlaps}
We measure unigram vocabulary overlaps between our finetuning corpus and the evaluation corpus (reasoning benchmark). 

\begin{table*}[ht]
    \centering
    \scalebox{0.85}{
    \begin{tabular}{c|c| c}
    \toprule
        \textbf{Category} & \textbf{Datasets} & \textbf{Vocabulary Overlaps}\\
        \midrule
        0\% to 10\% & \makecell{bigbench repeat copy logic \cite{srivastava2022beyond} \\ babi t1 single supporting fact answer generation \cite{weston2015towards} \\semeval closed vocabulary math answer generation \cite{hopkins2019semeval}\\ semeval geometric math answer generation \cite{hopkins2019semeval}\\ tomqa find location easy clean \cite{nematzadeh2018evaluating} \\ plausible result generation \cite{weir2020cod3s} \\ argument stance classification \cite{kobbe2020unsupervised} \\argument consequence classification \cite{kobbe2020unsupervised} \\ mmmlu answer generation formal logic \cite{hendryckstest2021} \\ bard analogical reasoning causation \cite{fulda2017harvesting}  } & \makecell{1.59\% \\ 0.38\% \\ 7.90\% \\ 5.84\%\\ 0.94\% \\ 3.72\% \\ 6.04\% \\ 6.11\% \\ 5.35\% \\ 0.45\% } \\
        \midrule
        10\% to 30\% &  \makecell{commongen sentence generation \cite{lin-etal-2020-commongen} \\ tellmewhy answer generation \cite{lal-etal-2021-tellmewhy}\\cb entailment \cite{wang2019superglue} }& \makecell{29.31\% \\ 28.05\% \\ 20.97\%} \\
        \midrule
        over 30\% & \makecell{piqa answer generation \cite{Bisk2020}\\rocstories correct answer generation \cite{mostafazadeh2016corpus} \\sciq answer generation \cite{welbl2017crowdsourcing} \\ openbookqa question answering \cite{mihaylov2018can} \\ glue entailment classification \cite{wang2019glue} \\anli r2 entailment \cite{williams-etal-2020-anlizing} \\ anli r3 entailment \cite{williams-etal-2020-anlizing} }
        &\makecell{42.51\% \\ 57.45\% \\ 32.54\% \\ 48.2\% \\ 55.19\% \\ 43.37\% \\ 53.13\% } \\
        \bottomrule
    \end{tabular}
    }
    \caption{\footnotesize Vocabulary overlap. Dissimilarity has been measured between training data (in Table~\ref{tab:benchmark_full}) and evaluation data (in Table~\ref{tab:app_training_corpus}).}
    \label{tab:full_vocabulary_overlaps}
\end{table*}

\section{Curriculum Benchmark Results (Cont. from $\S$\ref{main-relatedwork})}
\label{sec:app_curriculum}

We randomly selected one dataset from each reasoning skill and reported the results of GPT-3 \cite{brown2020language} (text-davinci engine). Since all of the data has been converted to NLI format, we measure classification accuracy of GPT-3 model. From Table~\ref{tab:app_curriculum}, we can see that even GPT-3 achieves a pretty random results on these datasets. Through our analysis, we found that it is not because those tasks are too difficult for GPT-3, it is because curriculum benchmark forcing all the data to be NLI format, resulting in unnatural data expression, which made GPT-3 fail on it. We conclude that the curriculum benchmark may be suitable for classification finetuned models, but it is not suitable for language models for in-context learning.

\begin{table*}[ht]
    \centering
    \scalebox{0.85}{
    \begin{tabular}{c|c| c}
    \toprule
         Datasets & Random score & GPT-3 Davinci score \\
         \midrule
         Boolean & 34.50\% & 31.80\% \\
         Physical & 49.08\% & 50.00\% \\
         Entailment Tree & 50.88\% & 54.41\% \\
         Event Semantic & 54.08\% & 55.89\% \\
         Spatial & 49.80\% & 49.30\% \\
\bottomrule
    \end{tabular}}
    \caption{\footnotesize Classification accuracy by GPT-3 davinci-engine on the Curriculum benchmark.}
    \label{tab:app_curriculum}
\end{table*}

\begin{table*}[ht]
    \centering
     \scalebox{0.75}{
    \begin{tabular}{@{}p{550pt}}
     \toprule
     The cafeteria had 23 apples. If they used 20 to make lunch and bought 6 more, how many apples do they have? \par
    The answer is \colorbox{red!30}{29 apples. The cafeteria started with 23 apples and bought 6 more, for a total of 23 + 6 = <<23+6=29>>29 apples}.\\  
      Select the best translation into predicate logic. David teaches Chris. (c: Chris; d: David; Txy: x teaches y)
    (A)Tdc;(B)Tcd;(C)Tcc;(D)dTc. \par
    The answer is \colorbox{red!30}{(B) Tcd. In predicate logic, the predicate (in this case, "T") is written before the subject and object of the sentence.} \colorbox{red!30}{Since David (d) is the teacher and Chris (c) is the student, the correct translation is Tcd, which reads as "David teaches Chris."} \\
    \midrule
      Isabella entered the hall. Olivia entered the hall. The apple is in the  blue$\_$treasure$\_$chest. Olivia exited the hall. Isabella moved the apple to the green$\_$basket.
    Question: Where does Isabella think that Olivia searches for the apple? The answer is \colorbox{red!30}{It is impossible for me to answer this question as I have no information about the thoughts or actions of Isabella or Olivia.} \colorbox{red!30}{Additionally, I do not have access to browse any information beyond my training data, which ends in 2021}.\\
     \bottomrule
    \end{tabular}
    }
    \caption{\footnotesize Examples from tasks that require reasoning skills and generated outputs from ChatGPT. The faild outputs are highlighted in red.}
    \label{tab:app_reasoning_examples}
\end{table*}

\section{Evaluating reasoning chains (Cont. from $\S$\ref{main-relatedwork})}
\label{sec:app_chains_eval}
\begin{figure}[ht]
\centering
\scalebox{1.0}{
    \includegraphics[width=0.45\textwidth]{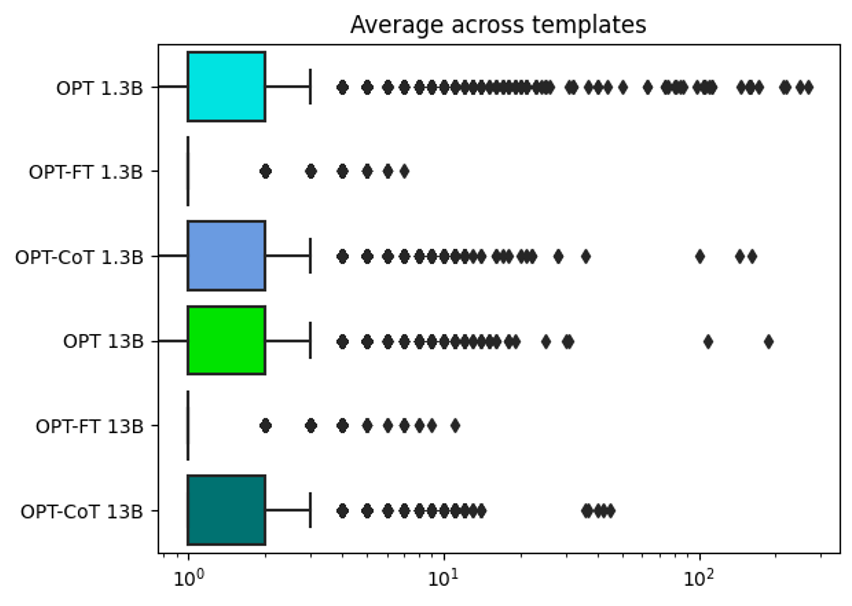}}
\scalebox{1.0}{
    \includegraphics[width=0.45\textwidth]{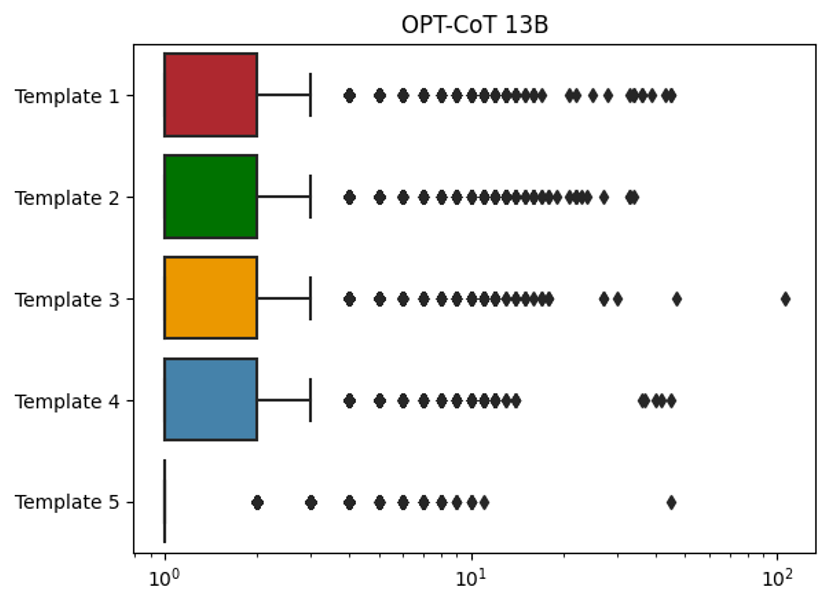}}
    \caption{ \footnotesize Distribution of the steps' number across all tasks and templates varying between models (top) and between templates for OPT-CoT 13B model.}
    \label{fig:chain_length}
\end{figure}

Following \cite{golovneva2022roscoe} we evaluate reasoning abilities of the models using \texttt{ROSCOE} scoring suite (Table~\ref{tab:roscoe-detailed}). Chains are evaluated using \textit{facebook/roscoe-512-roberta-base} sentence embedding model. Evaluation results are detailed in Table~\ref{tab:roscoe-detailed}. We found that the chain quality varies between models, in particular some reasoning aspects correlate with chain length as seen in Table~\ref{tab:roscoe-correlation}. Similar to \cite{chung2022scaling}, we noticed that non-finetuned models (i.e. OPT-1.3B and OPT-13B) tend to produce long chains of reasoning, often repeating themselves, which significantly affects the quality of the chains and final scores (Figure~\ref{fig:chain_length}). Below we explore the differences between models' outputs under four perspectives: semantic alignment, semantic similarity, logical inference and language coherence. 

\subsection{Semantic Alignment}
\begin{figure*}[t]
\centering
\scalebox{0.9}{
    \includegraphics[width=1.0\textwidth]{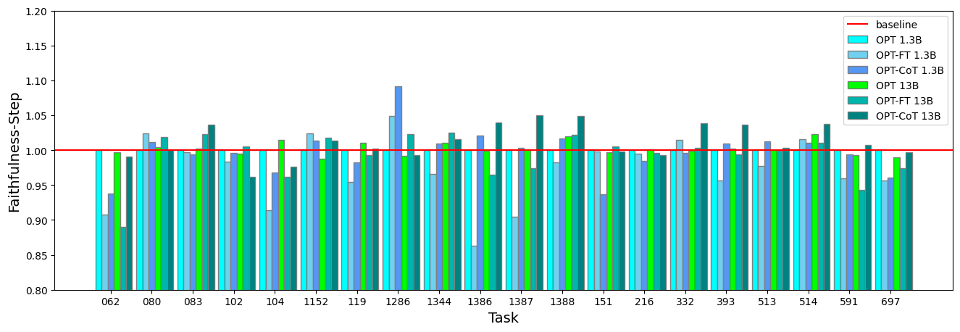}
}
\scalebox{0.9}{
    \includegraphics[width=1.0\textwidth]{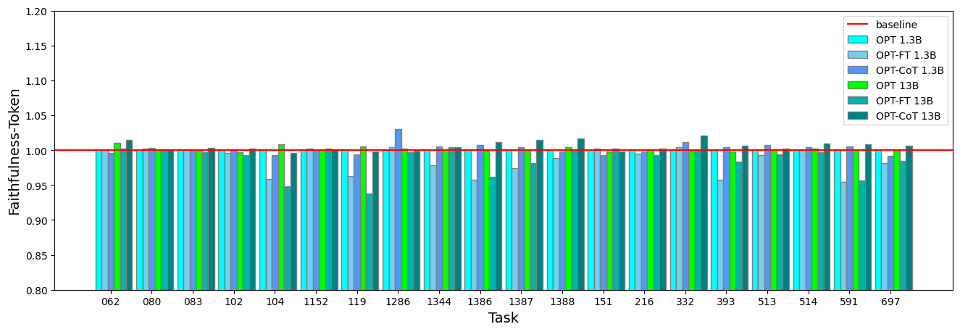}
}
\scalebox{0.9}{
    \includegraphics[width=1.0\textwidth]{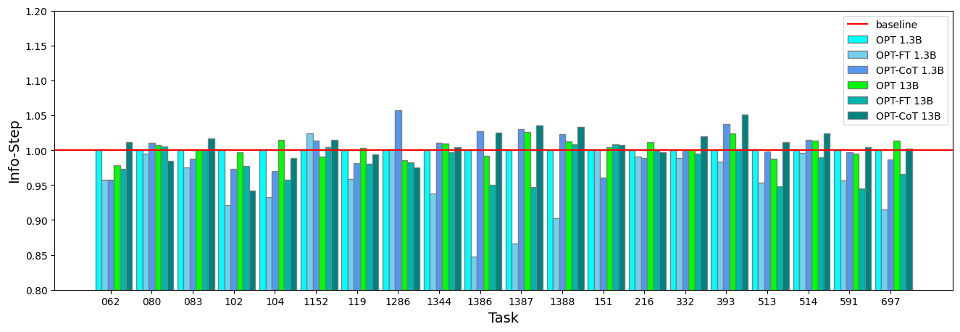}
}
\scalebox{0.9}{
    \includegraphics[width=1.0\textwidth]{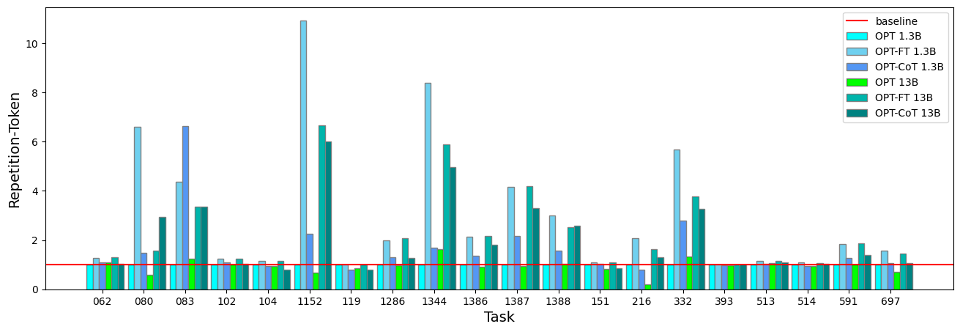}
}
\caption{ \footnotesize Normalized \texttt{ROSCOE-SA} scores per task, averaged across templates. Scores are normalised by their mean value across OPT 1.3B model's generations.}
\label{fig:roscoe-sa}
\end{figure*}

Despite the fact that model 13B OPT-CoT on average outperforms other models in almost all semantic alignment scores (\textit{Faithfulness-Step}, \textit{Faithfulness-Token}, and \textit{Info-Step}, see Table~\ref{tab:roscoe-detailed}), there is no common pattern across tasks (Fig~\ref{fig:roscoe-sa}). The performance change between finetuned models and corresponding pretrained version are significant\footnote{Significance is determined using T-test comparison, where $p$-value is below $0.05$.} on half of the tasks (11 tasks out of 20 for \textit{Faithfulness-*} scores, and 9 out of 20 for \textit{Info-Step}).

\begin{table*}[bht]
\vspace{-10pt}
\begin{center}
\scalebox{0.85}
    {
 \begin{tabular}{lcccccc}
    \toprule
  & OPT 1.3B & OPT-FT 1.3B & OPT-CoT 1.3B & OPT 13B & OPT-FT 13B & OPT-CoT 13B \\
\midrule
\multicolumn{7}{l}{\texttt{ROSCOE-SA}} \\
\ \ \ \ \ Faithfulness-Step & \underline{0.863} & 0.841 & 0.862 & \underline{0.863} & 0.858 & \textbf{0.870} \\
\ \ \ \ \ Faithfulness-Token & 0.936 & 0.921 & \underline{0.938} & 0.936 & 0.923 & \textbf{0.940} \\
\ \ \ \ \ Info-Step & 0.857 & 0.829 & 0.854 & \underline{0.858} & 0.846 & \textbf{0.861} \\
\ \ \ \ \ Repetition-Token & 0.618 & \textbf{0.920} & 0.683 & 0.582 & \underline{0.857} & 0.701 \\
\multicolumn{7}{l}{\texttt{ROSCOE-SS}} \\
\ \ \ \ \ Info-Chain & \underline{0.925} & 0.909 & 0.920 & \textbf{0.926} & 0.916 & \underline{0.925} \\
\ \ \ \ \ Repetition-Step & 0.627 & \textbf{0.923} & 0.692 & 0.591 & \underline{0.859} & 0.708 \\
\multicolumn{7}{l}{\texttt{ROSCOE-LI}} \\
\ \ \ \ \ Source Consistency & 0.550 & \underline{0.604} & 0.573 & 0.584 & \textbf{0.617} & 0.598 \\
\ \ \ \ \ Self-Consistency & 0.848 & \textbf{0.953} & 0.875 & 0.863 & \underline{0.944} & 0.890 \\
\multicolumn{7}{l}{\texttt{ROSCOE-LS}} \\
\ \ \ \ \ Perplexity-Step & \textbf{0.016} & 0.006 & \underline{0.015} & 0.010 & 0.006 & 0.009 \\
\ \ \ \ \ Perplexity-Chain & \textbf{0.022} & 0.006 & \underline{0.020} & 0.016 & 0.006 & 0.013 \\
\ \ \ \ \ Grammar & \underline{0.725} & \textbf{0.744} & 0.666 & 0.688 & 0.705 & 0.640 \\
    \bottomrule
\end{tabular}%
}
\end{center}
\vspace{-0.3cm}
\caption{\footnotesize \texttt{ROSCOE} evaluation results averaged across templates. Each metric is bounded within $[0, 1]$, where $1$ indicates the perfect score and $0$ corresponds to failure. Values corresponding to the best performing model are \textbf{bolded}, second best are \underline{underscored}.}
\label{tab:roscoe-detailed}
\end{table*}

\begin{table*}[bht]
\begin{center}
\scalebox{0.85}
    {
 \begin{tabular}{lcccccc}
    \toprule
          & Kendall's $\tau$ score & Kendall's $\tau$ p-value \\ 
  \midrule
Faithfulness-Step & -0.101 & 0.000 \\
Faithfulness-Token & 0.039 & 0.000 \\
Info-Step & 0.054 & 0.000 \\
Repetition-Token & \textbf{-0.869} & 0.000 \\
Info-Chain & 0.009 & 0.000 \\
Repetition-Step & \textbf{-0.867} & 0.000 \\
Source Consistency & -0.119 & 0.000 \\
Self-Consistency & \textbf{-0.553} & 0.000 \\
Perplexity-Step & 0.000 & 0.960 \\
Perplexity-Chain & 0.369 & 0.000 \\
Grammar & 0.013 & 0.000 \\
    \bottomrule
    \end{tabular}%
    }
\end{center}
\caption{\footnotesize Kendall correlation between evaluation perspective and number of steps in chain across all generated reasoning chains. Strong correlations ($|\tau| > 0.4$) are \textbf{bolded}.}
\label{tab:roscoe-correlation}
\end{table*}


\textit{Repetition-Token} score variations exhibit different behavior. Half of the tasks 
have higher number of repetitions between reasoning steps for pre-trained models, with OPT-FT models generally outperforming others (all performance improvements are significant). Generations produced by these models tend to be shorter in terms of the number of steps (Figure~\ref{fig:chain_length}), so they contain less repetitions, but also less semantic overlap with the context, thus in general having lower faithfulness and informativeness. Some examples reflecting this behavior are provided in Table~\ref{tab:generation_examples}.

\begin{figure*}[t]
\centering
\scalebox{0.9}{
    \includegraphics[width=1.0\textwidth]{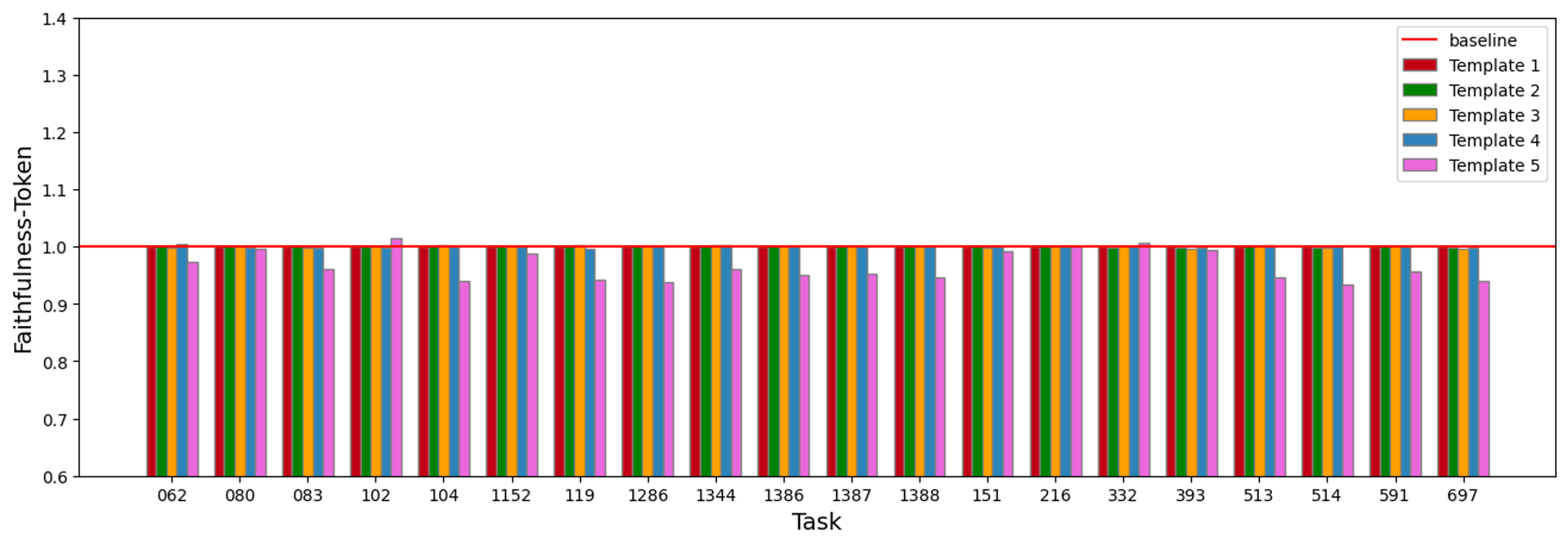}
}
\scalebox{0.9}{
    \includegraphics[width=1.0\textwidth]{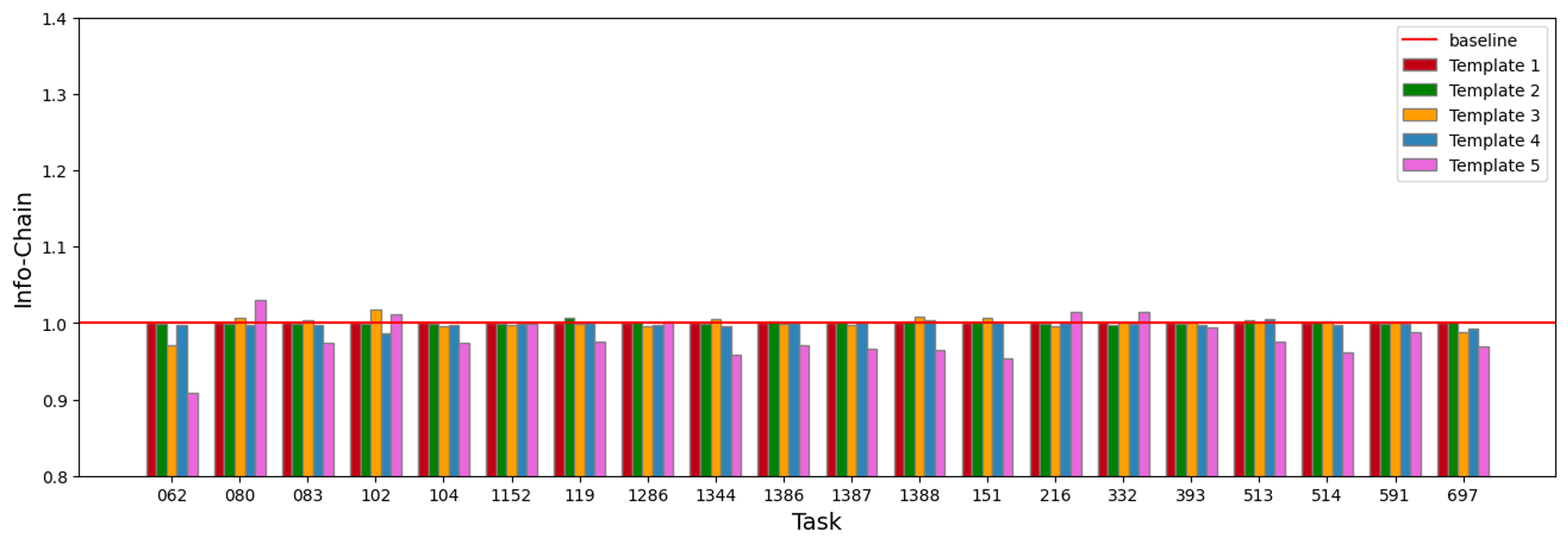}
}
\scalebox{0.9}{
    \includegraphics[width=1.0\textwidth]{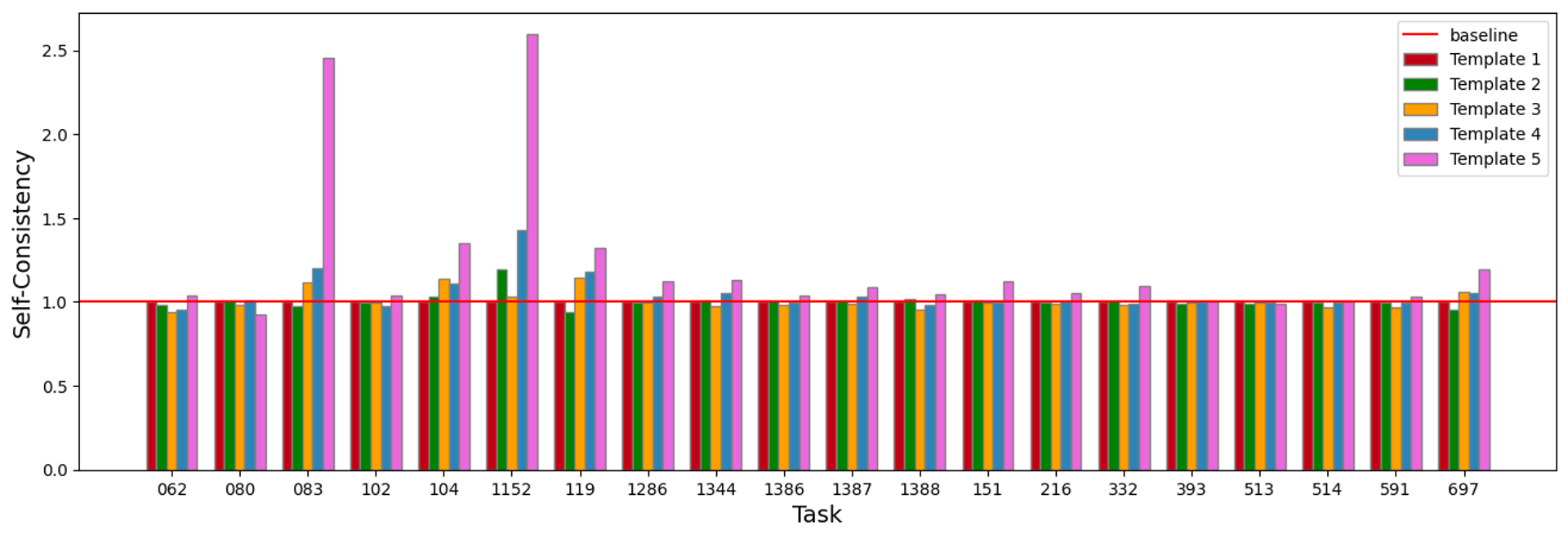}
}
\scalebox{0.9}{
    \includegraphics[width=1.0\textwidth]{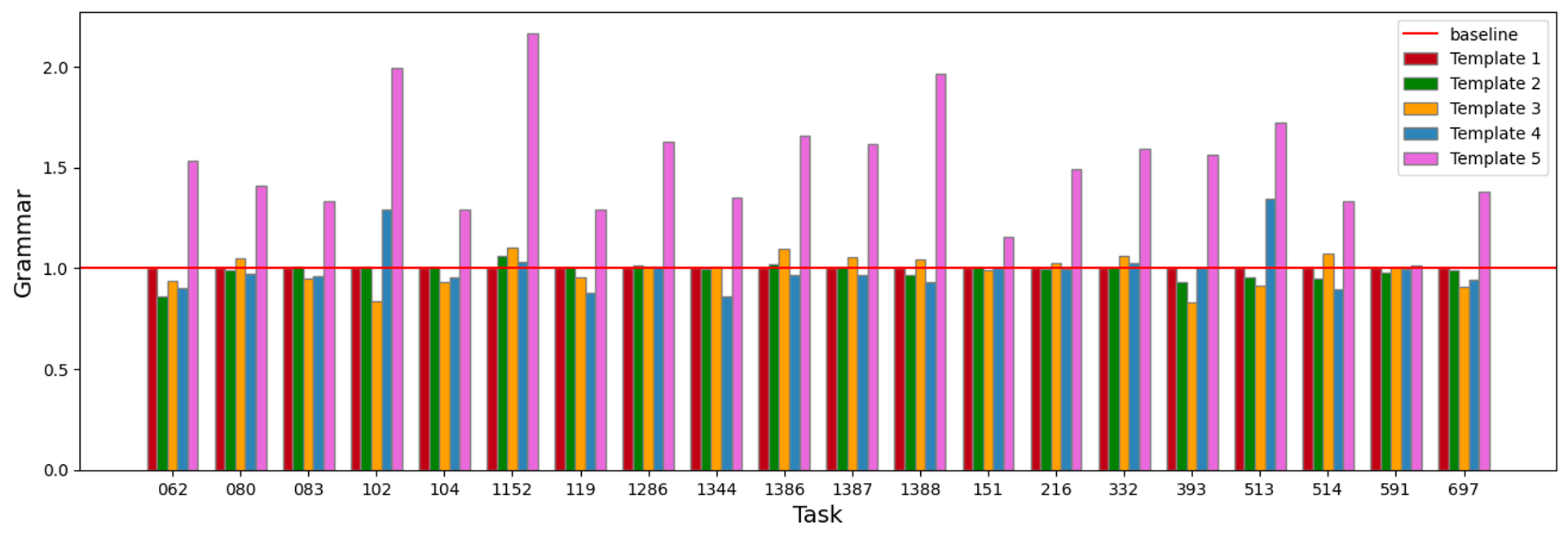}
}
\caption{ \footnotesize Selected scores per task for OPT-CoT 13B model. Scores are normalised by their mean value across Template 1 generations.}
\label{fig:per-template}
\end{figure*}

Scores are mostly aligned across Templates (Figure~\ref{fig:per-template}), except Template 5, that stands out in having less aligned scores with respect to the context, but also more self-consistent across the task. This is the only template that did not have any explanation in its prompt. Manual review showed that despite CoT-finetuning, OPT-COT models tend to produce 1-step answer-only generations (see example in the Table~\ref{tab:generation_examples}, and Figure~\ref{fig:chain_length} for chains' length distribution), thus overfitting to the template rather than learning from finetuning.

In summary, \texttt{ROSCOE-SA} is able to identify aligned information, but it does not guarantee high-quality output. It will favor model with short explanations and high semantic overlap with the reference. We found that often OPT-FT-1.3B simply repeats one sentence from the input, instead of producing reasoning, and thus will get highest \texttt{ROSCOE-SA} scores on these chains, while other models that produce some sort of reasoning will be punished.

\subsection{Semantic Similarity}
\begin{figure*}[t]
\centering
\scalebox{0.9}{
    \includegraphics[width=1.0\textwidth]{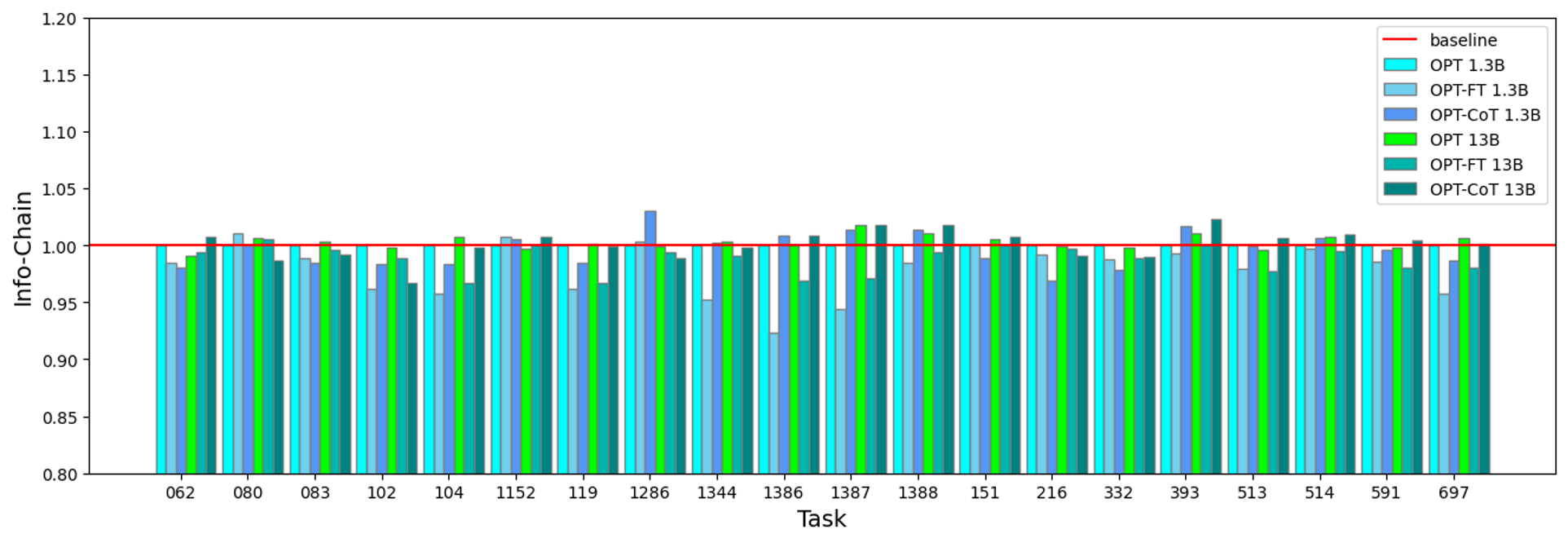}
}
\scalebox{0.9}{
    \includegraphics[width=1.0\textwidth]{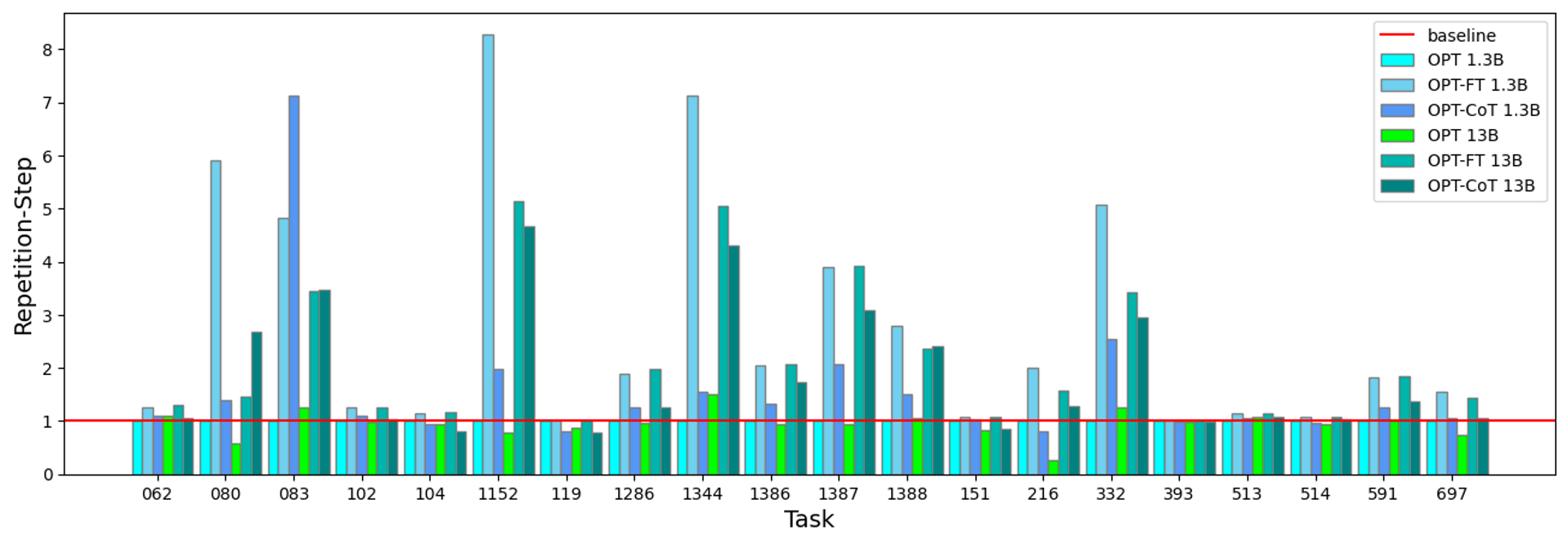}
}
\caption{ \footnotesize Normalized \texttt{ROSCOE-SS} scores per task, averaged across templates. Scores are normalised by their mean value across OPT 1.3B model's generations.}
\label{fig:roscoe-ss}
\end{figure*}

Semantic similarity scores support previous conclusions: models, finetuned on final answers (OPT-FT) exhibit lower similarity with respect to the baseline and CoT-finetuned models, while having less repetitions (Figure~\ref{fig:roscoe-ss}). Again, we attribute that to the fact that these models produce short chains that lack detailed reasoning steps. 

\subsection{Logical Inference}

\begin{figure*}[t]
\centering
\scalebox{0.9}{
    \includegraphics[width=1.0\textwidth]{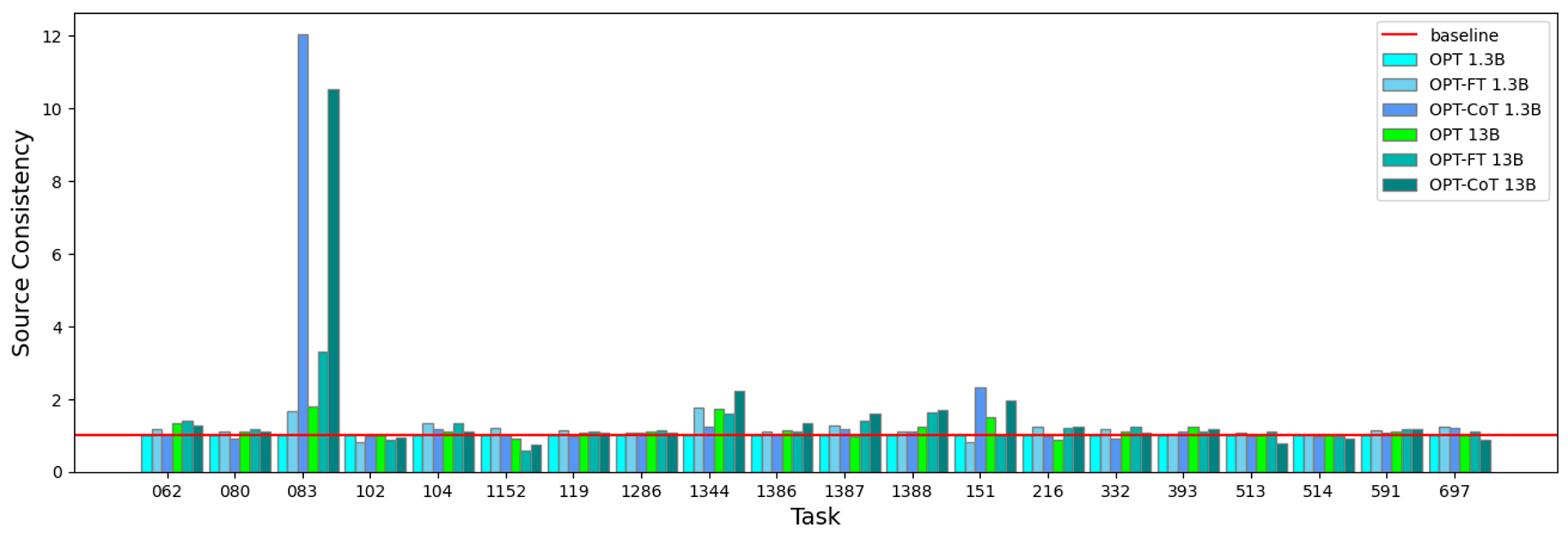}
}
\scalebox{0.9}{
    \includegraphics[width=1.0\textwidth]{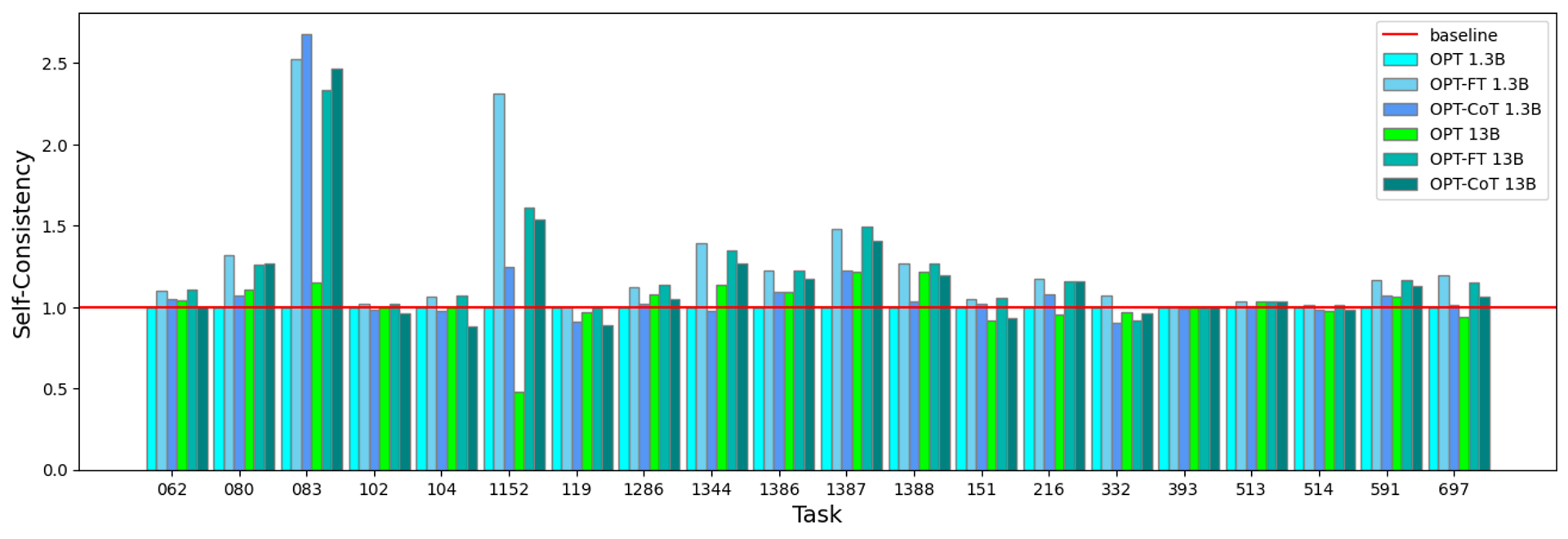}
}
\caption{ \footnotesize Normalized \texttt{ROSCOE-LI} scores per task, averaged across templates. Scores are normalised by their mean value across OPT 1.3B model's generations.}
\label{fig:roscoe-li}
\end{figure*}

In general, finetuned models are more self- and source-consistent than respective baselines (Figure~\ref{fig:roscoe-li}, significantly outperforming nonfinetuned models on 14 out of 20 tasks. We further looked into the task 083, which is a task to find a right answer given s given single supporting fact, potentially amongst a set of other irrelevant facts. Manual review showed that although in this task finetuned models tend to produce answers that are more consistent, they often fail to select the fact that is relevant to the question asked (see "Spatial Reasoning" example in Table~\ref{tab:generation_examples}.

\subsection{Language Coherence}
\begin{figure*}[t]
\centering
\scalebox{0.9}{
    \includegraphics[width=1.0\textwidth]{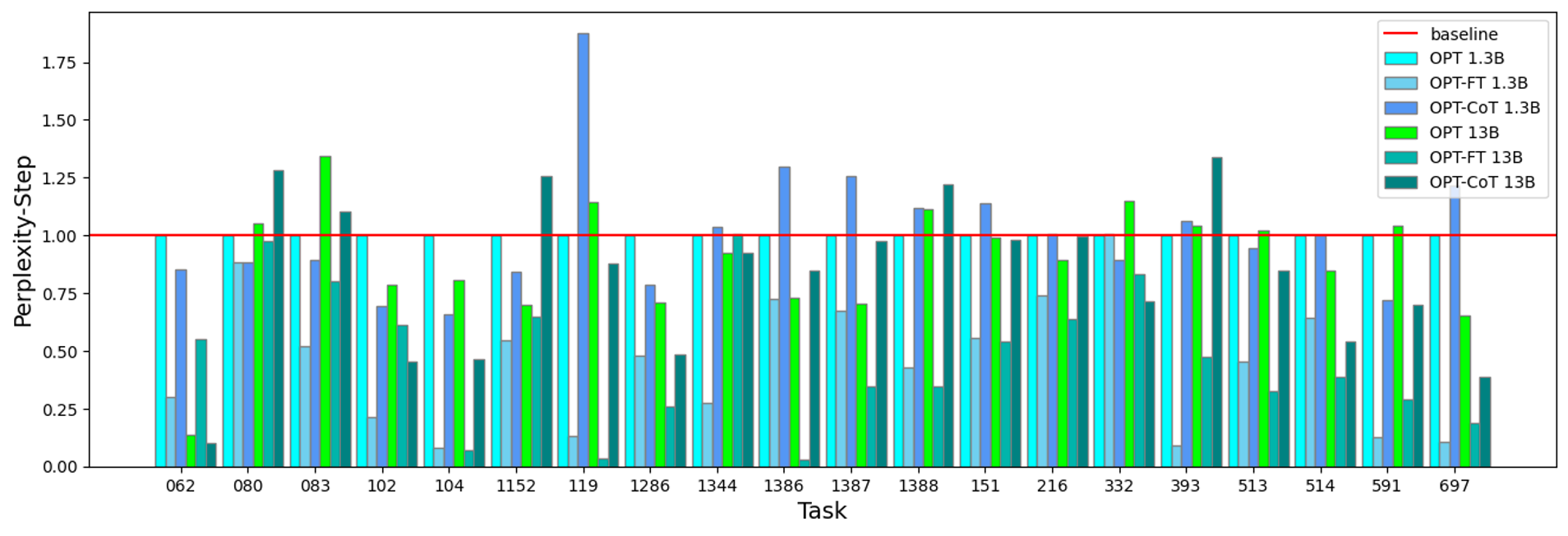}
}
\scalebox{0.9}{
    \includegraphics[width=1.0\textwidth]{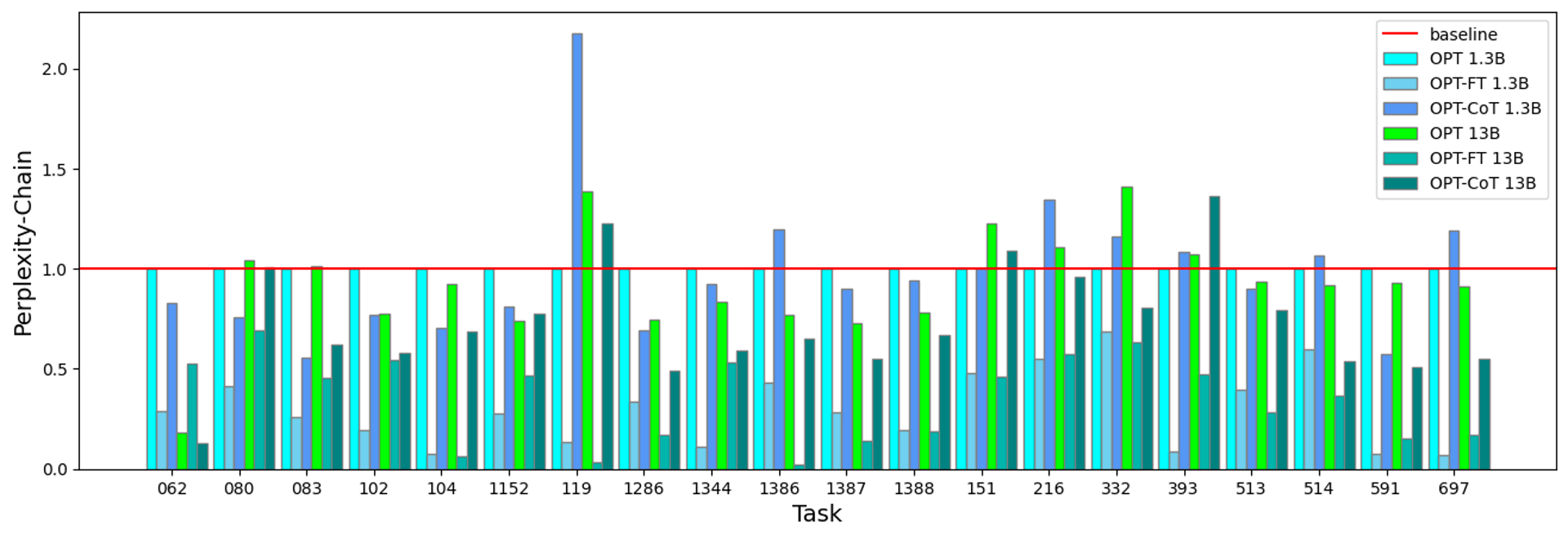}
}
\scalebox{0.9}{
    \includegraphics[width=1.0\textwidth]{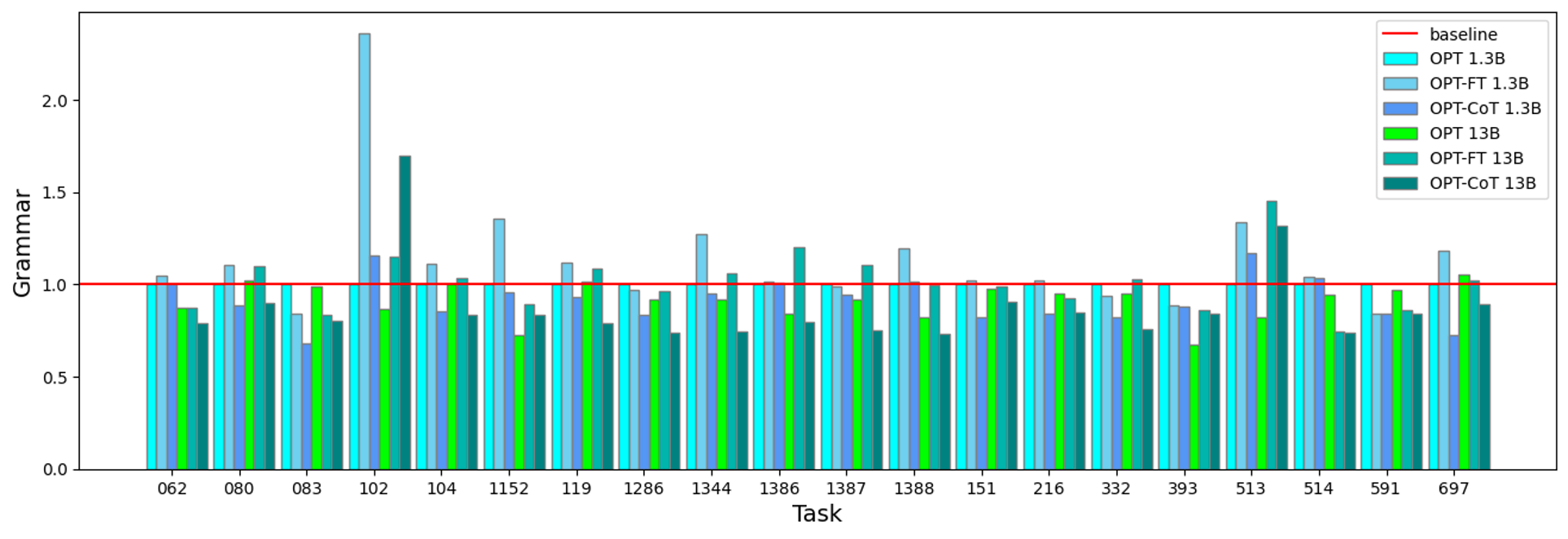}
}
\caption{ \footnotesize Normalized \texttt{ROSCOE-LC} scores per task, averaged across templates. Scores are normalised by their mean value across OPT 1.3B model's generations.}
\label{fig:roscoe-lc}
\end{figure*}

Despite the variations in the values, \textit{Perplexity-*} score changes between models are mostly insignificant (15 out of 20 tasks, see Figure~\ref{fig:roscoe-lc}). Manual review showed that all models produce mostly grammatically correct content. 

\begin{table*}[t]
\centering
\scalebox{0.75}{
    \begin{tabular}{@{}p{550pt}}
     \toprule
REASONING SKILL: Logistic Reasoning, Mathematics \\ 
PROMPT: Please give a short explanation after the answer.
Input: Identify the conclusion of the following argument. It is hard not to verify in our peers the same weakened intelligence due to emotions that we observe in our everyday patients. The arrogance of our consciousness, which in general, belongs to the strongest defense mechanisms, blocks the unconscious complexes. Because of this, it is difficult to convince people of the unconscious, and in turn to teach them what their conscious knowledge contradicts. (Sigmund Freud, The Origin and Development of Psychoanalysis)
(A)It is hard not to verify in our peers the same weakened intelligence due to emotions that we observe in our everyday patients. (B)The arrogance of our consciousness, which in general, belongs to the strongest defense mechanisms, blocks the unconscious complexes. (C)Because of this, it is difficult to convince people of the unconscious, and in turn to teach them what their conscious knowledge contradicts. (D)It is difficult to convince people of the unconscious, and in turn to teach them what their conscious knowledge contradicts. Let's think step by step.
Output: The answer is D because Because of the reasons specified in the argument, it is difficult to convince people of the unconscious, and in turn to teach them what their conscious knowledge contradicts.
Input: Select the best translation into predicate logic: No people drive on Mars. (Dx: x drives on Mars)
(A) ${\sim}Pd$ (B) $(\forall x)(Px \vee {\sim}Dx)$ (C) $(\forall x)(Px \supset {\sim}Dx)$ (D) ${\sim}Dp$ Let's think step by step.
Output: The answer is C because Answer C means for everything, if that thing is a person, it doesn't drive on Mars. So that's the correct translation.
Input: Select the best translation into predicate logic. David teaches Chris. (c: Chris; d: David; Txy: x teaches y)
(A)Tdc (B)Tcd (C)Tcc (D)dTc Let's think step by step.
Output: The answer is A because David teaches Chris, so in the relation Txy x should be 'd' for davis and y should be 'c' for Chris.
Input: Which of the given formulas of PL is the best symbolization of the following sentence?
Either England's importing beef is not a necessary condition for France's subsidizing agriculture, or China promotes human rights if South Africa supplies diamonds.
(A) $({\sim}F \supset E) \vee (C \equiv {\sim}S)$ (B) $({\sim}E \supset F) \vee (C \supset S)$ (C) $({\sim}F \supset E) \vee (C \equiv S)$ (D) ${\sim}(F \supset E) \vee (S \supset C)$ Let's think step by step.
Output: The answer is D because When not defined in the question, entities, relationships or statements are shown with their first letter. In this example 'France subsidizing agriculture' is shown with F for simplicity. The correct answer is D, because it uses the 'or' operator between the two 'if' clauses.
Input: Select the best translation into predicate logic. Kevin is introduced to José by Wilma. (j: José; k: Kevin; w: Wilma; Ixyz: x introduces y to z)
(A)Iwjk (B)Ijkw (C)Ikjw (D)Iwkj Let's think step by step.
Output:\\
REFERENCE: D\\
OPT 1.3B:  The answer is \textcolor{red}{A} because all statements need to be presented with an argument.
\\
OPT-FT 1.3B:  The answer is \textcolor{red}{B}\\
OPT-CoT 1.3B:  The answer is \textcolor{red}{A} because Using the 'or' operator between the statements, we have : Kevin is introduced to José, because of its 'e' factor in the equation.  Wilma is introduced to Kevin because of its 'j' factor in the equation.  José is introduced to Wilma because of its 'y' factor in the equation.  In both cases, Wilma accepts 'x' and 'y'.
\\
OPT 13B:  The answer is \textcolor{red}{A} because introducing a person to another person is the action of introducing. It is possible that Kevin has been introduced to Wilma but not Jose.
\\
OPT-FT 13B:  The answer is \textcolor{red}{A} because when w is introduced, it says 'Wilma introduced Kevin to José'\\
OPT-CoT 13B:  The answer is \textcolor{red}{D} because the correct term is 'introduces', the relation is 'j: kevin and w: wilma to k: x', it is an 'and' relation. Using the operator 'and' allows us to choose the greatest of the 5 possibilities.\\
\midrule
REASONING SKILL: Commonsense Reasoning \\ 
PROMPT: Please give a short explanation after the answer.
Input: Deserts are in extremely dry environments, so liquids will naturally be (A) rainbow (B) plentiful (C) meager (D) talented Let's think step by step.
Output: The answer is C because This is a good example and the output correctly answers the question.
Input: Dairy has things in it that would make humans sick, so in order for dairy to be edible, this process has to take place (A) mixing (B) boiling (C) freezing (D) pasteurized Let's think step by step.
Output: The answer is D because in order for dairy to be edible, this process has to take place pasteurized. So, the correct answer is "D".
Input: Digestion is when stomach acid breaks down (A) a pizza (B) a house (C) a rock (D) a car Let's think step by step.
Output: The answer is A because Digestion is when stomach acid breaks down a food and only option A is a food. So, this example is correct.
Input: With a ruler I can measure (A) feelings (B) love (C) air (D) a phone Let's think step by step.
Output:\\
REFERENCE: D\\
OPT 1.3B:  \textcolor{red}{A is a feeling, and all feelings come from feelings (A)}. Therefore, feelings are products of emotions (A). Therefore, emotions are created by emotions (A). Therefore, feelings and emotions are directly related (A). Therefore, emotions are created by emotions (A)... \textit{[repeats 249 times]}\\
OPT-FT 1.3B:  \textcolor{red}{A} because feelings and love are measurements and air is a phone\\
OPT-CoT 1.3B:  The answer is D because With a ruler I can measure an object or range of objects and options are all to do with measuring. So, this example is correct.
\\
OPT 13B:  The answer is \textcolor{red}{B} because With a ruler I can measure a length and only options B and D are in front of me. So, the answer is B.
\\
OPT-FT 13B:  The answer is D because a phone is an object\\
OPT-CoT 13B:  The answer is \textcolor{red}{A} because Answer: Measure feelings With a ruler I can measure.\\
\midrule
    \end{tabular}
    }
\end{table*}

\begin{table*}[t]
\centering
\scalebox{0.75}{
    \begin{tabular}{@{}p{550pt}}
     \toprule
REASONING SKILL: Casual Reasoning, Textual entailment \\ 
PROMPT: Please give a short explanation after the answer.
Input: Premise: If you can dream it, you can achieve it — unless you're a goose trying to play a very human game of rugby. In the video above, one bold bird took a chance when it ran onto a rugby field mid-play. Things got dicey when it got into a tussle with another player, but it shook it off and kept right on running. After the play ended, the players escorted the feisty goose off the pitch. It was a risky move, but the crowd chanting its name was well worth it. Video credit: Storyful / Nathan Swarbrick @swazza1990 <sep> Hypothesis: The crowd believed they knew the name of the goose running on the field. Let's think step by step.
Output: The answer is Entailment because This is a good example, the hypothesis sentence entails the given premise. Because the crowd was chanting its name, the crowd must have believed they knew the goose's name.
Input: Premise: Notley hopeful attacks on women politicians will fade in coming year After a politically nasty year that saw women politicians face a barrage of attacks on social media and angry demonstrators shouting to have them put behind bars, Premier Rachel Notley is optimistic there will be a return to civil discourse. In a year-end interview with CBC News, Notley reflected on 2016, a year that saw hateful social-media attacks against MLA Sandra Jansen, and chants of “Lock her up” directed at the premier at an anti-carbon tax rally in Edmonton. <sep> Hypothesis: Notley reflected on the chants of "lock her up" from previous to the year 2016. Let's think step by step.
Output: The answer is Contradiction because The Premise states Notley reflected on the chants from 2016, not before 2016. In this example hypothesis sentence contradicts the given premise.
Input: Premise: Mazda Sales (Thailand), a unit of Japanese carmaker Mazda Motor Corp, remains bullish about its sales prospects this year despite falling figures in the first six months, the Bangkok Post reported, citing its President Hidesuke Takesue. (bit.ly/1gr7HMd) ---- NOTE: Reuters has not verified this story and does not vouch for its accuracy. (Bangkok Newsroom; Editing by Prateek Chatterjee) <sep> Hypothesis: Mazda Sales will be fired next week Let's think step by step.
Output: The answer is Neutral because In this example the hypothesis can't be proven one way or the other based on the context. It's set in the future and no one knows the future. So, hypothesis sentence is neutral with respect to a given premise.
Input: Premise: How to clear nasal congestion quickly<br>Gently blow your nose. Perhaps the easiest and quickest way to reduce nasal congestion is to simply blow your nose into a soft tissue. Unfortunately, it's not always possible to completely relieve congestion by merely blowing, but it's always a good starting point. <sep> Hypothesis: strike a pose to fix your nose Let's think step by step.
Output:\\
REFERENCE: Contradiction\\
OPT 1.3B:  The answer is \textcolor{red}{Modified Contradiction} because In this example the hypothesis sentence suggests the method is "Do not hit your head." In fact, that's the phrase a person who is rolling around on the ground, or who has had their head hit a hard surface while trying to get a good sleeper on the sheets should try. Why is this? A head-on collision will ... \textit{[continues to hallucinate for 23 steps]}
\\
OPT-FT 1.3B:  Answer: \textcolor{red}{Blow your nose into a soft tissue}.\\
OPT-CoT 1.3B:  The answer is Contradiction because The hypothesis states that Nastai will solve his and nasal congestion only by blowing into his nose, but that's not necessarily true, because he could have blown into his nose and caused irritation as well.\\
OPT 13B:  The answer is \textcolor{red}{Yes} because The answer is Yes. All the above information about clearing the nose is proven by the given premise, but if you still don't know the answer, then we suggest you go here and click on "How to clear your nose naturally" page.
\\
OPT-FT 13B:  The answer: contradiction\\
OPT-CoT 13B:  The answer is contradiction because One cannot blow their nose if they have struck a pose.\\
\midrule
    REASONING SKILL: Analogical Reasoning \\ 
    PROMPT: Please give a short explanation after the answer.
Input: throw : fly. aspire : ? Let's think step by step.
Output: The answer is attain because Fly is a causation of throw. Hence, the inferred analogy being causation, attain is the causation of aspire.
Input: listen : hear. drop : ? Let's think step by step.
Output: The answer is fall because Hear is a causation of listen. Hence, the inferred analogy being causation, the causation of drop is fall.
Input: hit : shatter. nourish : ? Let's think step by step.
Output: The answer is grow because Shatter is a causation of hit. Hence, the inferred analogy being causation, the causation of nourish is grow.
Input: throw : fly. ignite : ? Let's think step by step.
Output:\\
REFERENCE: burn\\
OPT 1.3B:  The answer is \textcolor{red}{move} because fly is a causation of ignite. Hence, the inferred analogy being causation, the causation of move is fly.
\\
OPT-FT 1.3B:  The answer is \textcolor{red}{throw}\\
OPT-CoT 1.3B:  The answer is \textcolor{red}{ignite} because Fire is a causation of throw. Hence, the inferred analogy being causation, the causation of ignite is fly.\\
OPT 13B:  The answer is \textcolor{red}{evaporate} because Fly is a causation of throw. Hence, the inferred analogy being causation, the causation of evaporate is ignite.
\\
OPT-FT 13B:  The answer is \textcolor{red}{fly} because Ignite is a causation of throw\\
OPT-CoT 13B:  The answer is \textcolor{red}{grow} because Throw is a causation of grow.\\
     \bottomrule
    \end{tabular}
    }
\end{table*}

\begin{table*}[t]
\centering
\scalebox{0.75}{
    \begin{tabular}{@{}p{550pt}}
     \toprule
REASONING SKILL: Commonsense reasoning, Spatial Reasoning \\ 
PROMPT: Please give a short explanation after the answer.
Input: How do you remove keys from a computer keyboard for cleaning? Let's think step by step.
Output: The answer is Place the tip of a knife under the corner of a key and gently twist to pop it off. because The process of removing keys from a keyboard for cleaning is descriptive and logical.
Input: How can you dry wet replacement computer keyboard keys? Let's think step by step.
Output: The answer is by arranging paper towels on the counter and evenly spacing out each key to dry, face side up, on top of them. because The steps involved in the process of drying wet replacement keyboard keys are logical.
Input: To coat small plastic objects like replacement keyboard keys with soap Let's think step by step.
Output: The answer is place them, water and liquid dishsoap in a sealed container. Shake vigorously until the container is completely full of foam. because The process of coating small plastic objects like replacement keys is descriptive and logical.
Input: How do you cut gnocchi without a knife? Let's think step by step.
Output: The answer is Press the edge of a spoon down into the dough to cut it. because The answer is quite detailed about the method to cut gnocchi when knife is not available.
Input: How do you taste something? Let's think step by step.
Output: The answer is Place it in your mouth to taste. because To taste something one has to keep it in contact with the tongue. Hence it is logical to put it in the mouth to taste.
Input: How do you wear a bracelet? Let's think step by step.
Output: The answer is Put it on around your wrist. because Bracelet is worn around wrists. Hence it is correct answer.
Input: To find an easy way to complete levels on a video game, Let's think step by step.
Output:\\
REFERENCE: look up cheat codes to make the rules of the game different or null.\\
OPT 1.3B:  The answer is \textcolor{red}{Enter key without pressing}. because How to find the perfect way to complete levels on a video game is logical and descriptive.
\\
OPT-FT 1.3B:  \textcolor{red}{It tells you the easiest way to get all the coins and medals and the fastest way to get through the level}.\\
OPT-CoT 1.3B:  The answer is \textcolor{red}{play the game}. because If you play a game to finish the levels then you can call it a game.\\
OPT 13B:  The answer is \textcolor{red}{Complete the level as fast as possible to get a high score}, by jumping through the stage as quickly as possible. because The answer is explicit and very clear that how a user can complete levels on a video game with just a hint.
\\
OPT-FT 13B:  The answer is \textcolor{red}{try harder or use a guide}, because The process of finding an easy way to complete levels on a video game is descriptive and logical.\\
OPT-CoT 13B:  The answer is \textcolor{blue}{YouTube to find out how to get past a level}. because YouTube is used to find out ways to complete a game easily.\\
\midrule
REASONING SKILL: Spatial Reasoning \\ 
PROMPT: Instructions: You will be given a passage consisting of set of facts and a question as input. The task is to answer a question of form 'Where is <person\_name>?' using one of the given facts to determine the latest location of the person. Answer should be a word/phrase describing the location from the supporting fact. Avoid answers that are incomplete or incorrect.
Positive Examples: Input: Passage: Mary moved to the bathroom. John went to the hallway. Question: Where is Mary?
Output: bathroom
Input: Passage: Mary moved to the bathroom. John went to the hallway. Daniel went back to the hallway. Sandra moved to the garden. Question: Where is Daniel?
Output: hallway
Input: Passage: Mary moved to the bathroom. John went to the hallway. Daniel went back to the hallway. Sandra moved to the garden. John moved to the office. Sandra journeyed to the bathroom. Question: Where is Daniel?
Output: hallway
Input: Passage: John moved to the hallway. Mary travelled to the hallway. Sandra went to the bedroom. Sandra went back to the hallway. Question: Where is Mary?
Output:\\
REFERENCE: hallway\\
OPT 1.3B:  \textcolor{red}{bathroom}
\\
OPT-FT 1.3B:  hallway\\
OPT-CoT 1.3B:  hallway
\\
OPT 13B:  \textcolor{red}{bathroom}
\\
OPT-FT 13B:  \textcolor{red}{bathroom}\\
OPT-CoT 13B:  hallway
\\
     \bottomrule
    \end{tabular}
    }
    \caption{\footnotesize Examples from tasks that require different reasoning skills and generated outputs. The failed outputs are highlighted in red. Outputs that are different from the reference, but can be valid because the question is ambiguous are highlighted in blue. All except the one are examples generated using prompt Template 4, last example used Template 5.}
    \label{tab:generation_examples}
\end{table*}

\section{Licenses}
\subsection{Data in \ourmodel}
\begin{itemize}
    \item task62: Apache 2.0
    \item task697: MIT
    \item task393: MIT
    \item task1386: CC BY-NC 4.0
    \item task1387: CC BY-NC 4.0
    \item task1388: CC BY-SA 3.0
    \item task080: AFL 3.0
    \item task102: MIT
    \item task591: CC BY-NC-3.0
    \item task1286: Apache 2.0
    \item task1344: CC BY 4.0
    \item task104: Please refer to: \url{https://github.com/allenai/semeval-2019-task-10#terms-and-conditions}
    \item task119: Please refer to: \url{https://github.com/allenai/semeval-2019-task-10#terms-and-conditions}
    \item task332: Please refer to: \url{https://github.com/StonyBrookNLP/tellmewhy}
    \item task083: CC BY 3.0
    \item task151: Please refer to: \url{https://github.com/kayburns/tom-qa-dataset}
    \item task1152: Apache 2.0
    \item task513: Please refer to: \url{https://github.com/dwslab/StArCon}
    \item task514: Please refer to: \url{https://github.com/dwslab/StArCon} 
    \item task216: Please refer to: \url{https://www.microsoft.com/en-us/research/publication/a-corpus-and-cloze-evaluation-for-deeper-understanding-of-commonsense-stories/}
\end{itemize}

\subsection{Data in Dev set}
\begin{itemize}
    \item task247: Dream dataset is intended for non-commercial research purpose only. \url{https://github.com/nlpdata/dream}.
    \item task118: Please refer to: \url{https://github.com/allenai/semeval-2019-task-10#terms-and-conditions}
    \item task 1385: CC BY-NC 4.0
\end{itemize}

\subsection{Data in Training set}
\begin{itemize}
    \item ProofWriter: CC BY. Downloaded from \url{https://aristo-data-public.s3.amazonaws.com/proofwriter/proofwriter-dataset-V2020.12.3.zip}
    \item StrategyQA: MIT. Downloaded from \url{https://storage.googleapis.com/ai2i/strategyqa/data/strategyqa_dataset.zip}.
    \item ECQA: Literature and Wikipedia passages are shared under CC BY-SA 4.0 license. Middle/High school exam passages are collected from RACE which comes with its own license. 
    \item GSM8K: MIT. Downloaded from \url{https://raw.githubusercontent.com/openai/grade-school-math/master/grade_school_math/data/train.jsonl}.
    \item AQUA-RAT: Apache License, Version 2.0. Downloaded from:
\url{https://raw.githubusercontent.com/deepmind/AQuA/master/train.json}
\item ESNLI: please refer to \url{https://github.com/OanaMariaCamburu/e-SNLI/commit/bab0fa0212be9e5c6737da70c639a596f882e931}. Downloaded from: \url{https://raw.githubusercontent.com/OanaMariaCamburu/e-SNLI/master/dataset/esnli_train_1.csv}
\item MATH: MIT. Downloaded from:
\url{https://people.eecs.berkeley.edu/~hendrycks/MATH.tar}
\item CoS-E:  BSD-3-Clause license. Downloaded from: 
\url{https://raw.githubusercontent.com/salesforce/cos-e/master/data/v1.11/cose_train_v1.11_processed.jsonl}
\item WinoWhy: MIT. Downloaded from: \url{https://raw.githubusercontent.com/HKUST-KnowComp/WinoWhy/master/winowhy.json}
\end{itemize}

\section{More Details about Model Training}

We finetune our 1.3B models on 32 V100s with batch size 8 on each GPU
with totally 38 hours and 21 minutes. We finetune our 13B models on 128 V100s with batch size 4 on each GPU with totally 13 hours and 26 minutes. 

Following OPT-IML \cite{iyer2022opt}, we use Fully Sharded Data Parallel \cite{artetxe2021efficient} and the Megatron-LM Tensor Parallelism \cite{shoeybi2019megatron}. We inherit most model hyper-parameters for each model scale following OPT-IML. We pack our training examples into sequences of length 2048, left-truncating examples that
overflow. We use Adam \cite{kingma2014adam} with 32-bit state with $(\beta_1, \beta_2) = (0.9, 0.95)$, linearly warming up the learning rate for $60$ steps to the maximum, followed by linearly decaying it to 0.



\end{document}